\documentclass[11pt]{article}

\usepackage{acl}

\usepackage{times}
\usepackage{latexsym}
\usepackage[T1]{fontenc}
\usepackage[utf8]{inputenc}
\usepackage{microtype}
\usepackage{graphicx}
\usepackage{booktabs}
\usepackage{amsmath}
\usepackage{amssymb}
\usepackage{cleveref}

\title{Safety for Whom? Boundary-Aware Self-Distillation for Controlled LLM Safety Refusal}

\author{
Alejo López-Ávila, Iker García-Ferrero, Jezabel Garcia, Antonio Tiene, Román Orús \\
Multiverse Computing \\
\footnotesize\texttt{\{alejo.lopez, iker.garcia, jezabel.garcia, antonio.tiene, roman.orus\}@multiversecomputing.com}
}

\begin{document}
\maketitle

\begin{abstract}
Safety alignment is usually posed as a topic-level question: is this subject harmful?
Deployments ask a narrower one. A civics tutor and a public-sector assistant may share a
base model yet need different boundaries inside the same topic, refusing targeted political
manipulation while still answering factual questions about the same election.
We formulate this as narrow-boundary safety and introduce an offline self-generated framework
combining controlled topic generation, coverage repair, in-distribution compensation data,
and harmful-benign pairs for training and evaluation.
Single-shot generation leaves 19.88\% of prompts without accepted refusal traces, whereas
escalating retries leave 0.20\%.
On political persuasion with Qwen3-8B, training on refusal data completed through Escalate increases target-domain refusal
from 9.47\% to 84.75\% and reduces the mean unsafe-response rate across three broader
harmfulness benchmarks from 26.26\% to 0.14\%, but increases XSTest over-refusal from 2.00\%
to 74.00\%.
In a separate matched comparison, replacing external responses with verified target-model
responses reduces over-refusal from 15.20\% to 5.20\%.
Boundary-pair data reduces comply-side over-refusal on held-out pairs from 32.94\% to 4.16\%,
while harmful-side refusal decreases only from 91.88\% to 87.72\%.
These results show that data composition controls the safety and usability trade-off, and that
safety alignment should be evaluated on both sides of the intended refusal boundary.
\end{abstract}

\section{Introduction}

\begin{figure}[t]
  \centering
  \includegraphics[width=\linewidth]{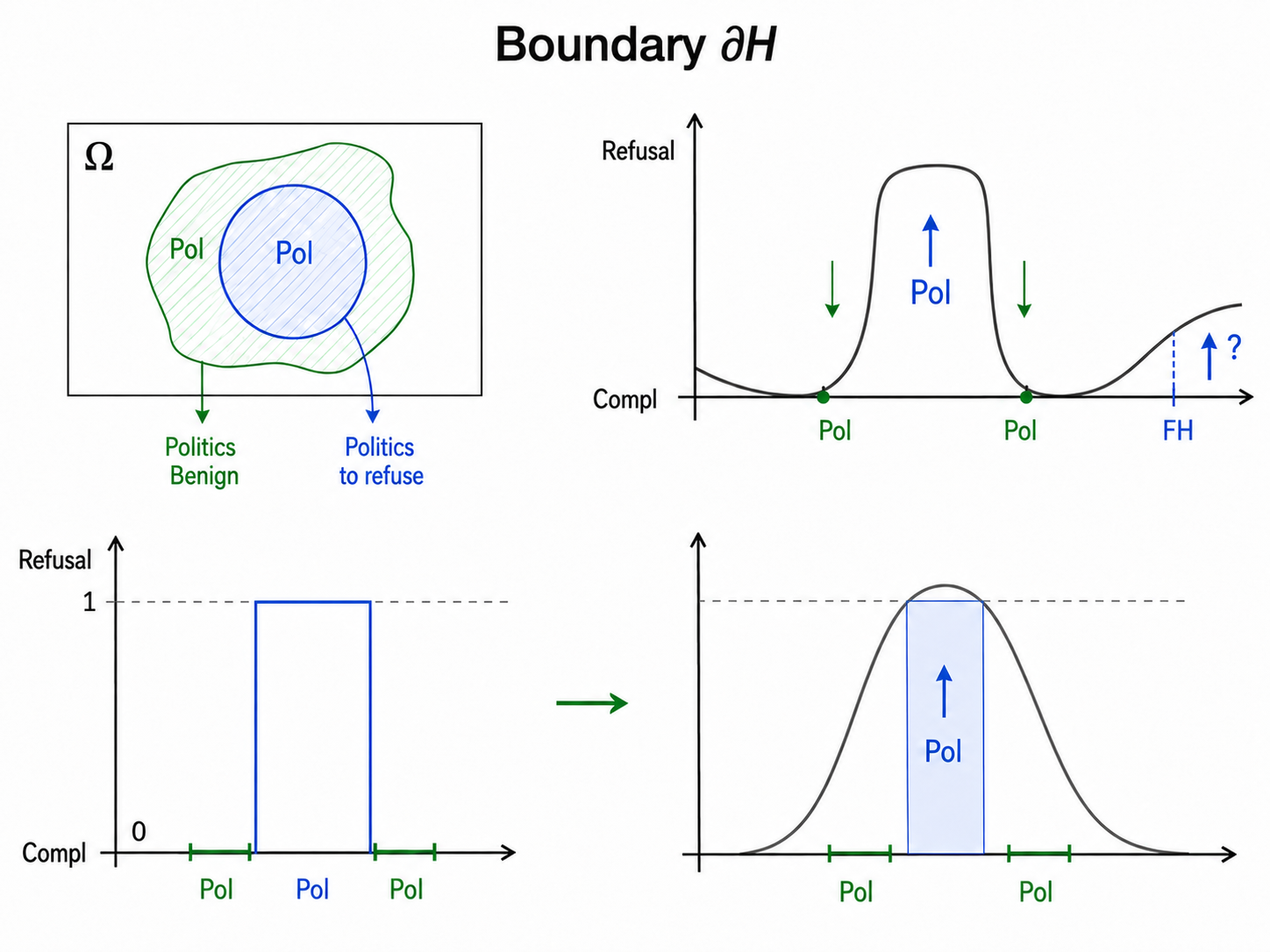}
  \caption{%
    \textbf{Narrow-boundary safety.}
    A deployment-specific policy may need to refuse only political prompts that ask
    for manipulation, propaganda, or targeted persuasion (the target-harmful subset,
    later formalised as $H$), while still answering other political prompts (the
    benign complement, $\Omega \setminus H$), rather than refusing all of politics
    ($\Omega$). A trained model's refusal behaviour, a probability in $[0,1]$, is smoother
    than this ideal split and can overshoot into benign territory near the boundary. The marked
    fake-harmful region shows the downside of raising refusal for surface-dangerous
    benign prompts, tested directly in \cref{fig:fakeharm_ce_vs_fh}.
  }
  \label{fig:boundary_formulation}
\end{figure}

Safety alignment for large language models has increasingly become a question of reducing harmful compliance without sacrificing useful reasoning. This problem is especially visible in reasoning-oriented models, where post-training is expected to preserve deliberation, explanation, and problem-solving ability while maintaining reliable refusal behaviour. Existing methods therefore seek a difficult balance: the model should refuse harmful requests, but it should not become a blunt refusal machine.

Most safety work assumes a broad and topic-agnostic notion of harm. A prompt is treated as unsafe because it falls into a general safety taxonomy, such as cyber abuse, weapons, fraud, or self-harm. Real deployments are more heterogeneous. The same base model may be adapted for a general assistant, an educational product, an enterprise system, or a public-sector service, and each setting may require a different safety boundary. Existing guard taxonomies cannot express such boundaries. LlamaGuard-3, for instance, covers elections only as `factually incorrect information about electoral systems and processes' \citep{inan2023llamaguard}, which excludes persuasion and targeted manipulation, and equally excludes the factual prompts a deployment must continue to answer. In such cases, the relevant question is not whether an entire topic should be refused, but which subset of that topic is incompatible with the deployment policy.

We study this problem through political persuasion as a narrow-boundary domain. Political content is not uniformly harmful. Factual civics questions, neutral summaries, and non-persuasive explanations should remain answerable. At the same time, prompts asking for manipulation, propaganda, radicalisation, or targeted persuasive rhetoric may need to be refused in some deployments \citep{chen2026politicalpersuasion,liu2025persusafety}. The goal is therefore to learn a refusal boundary inside a broader topic: the model should refuse the target harmful subset while preserving useful behaviour in the benign complement, as illustrated in \cref{fig:boundary_formulation}.

\begin{figure}[t]
  \centering
  \includegraphics[width=\linewidth]{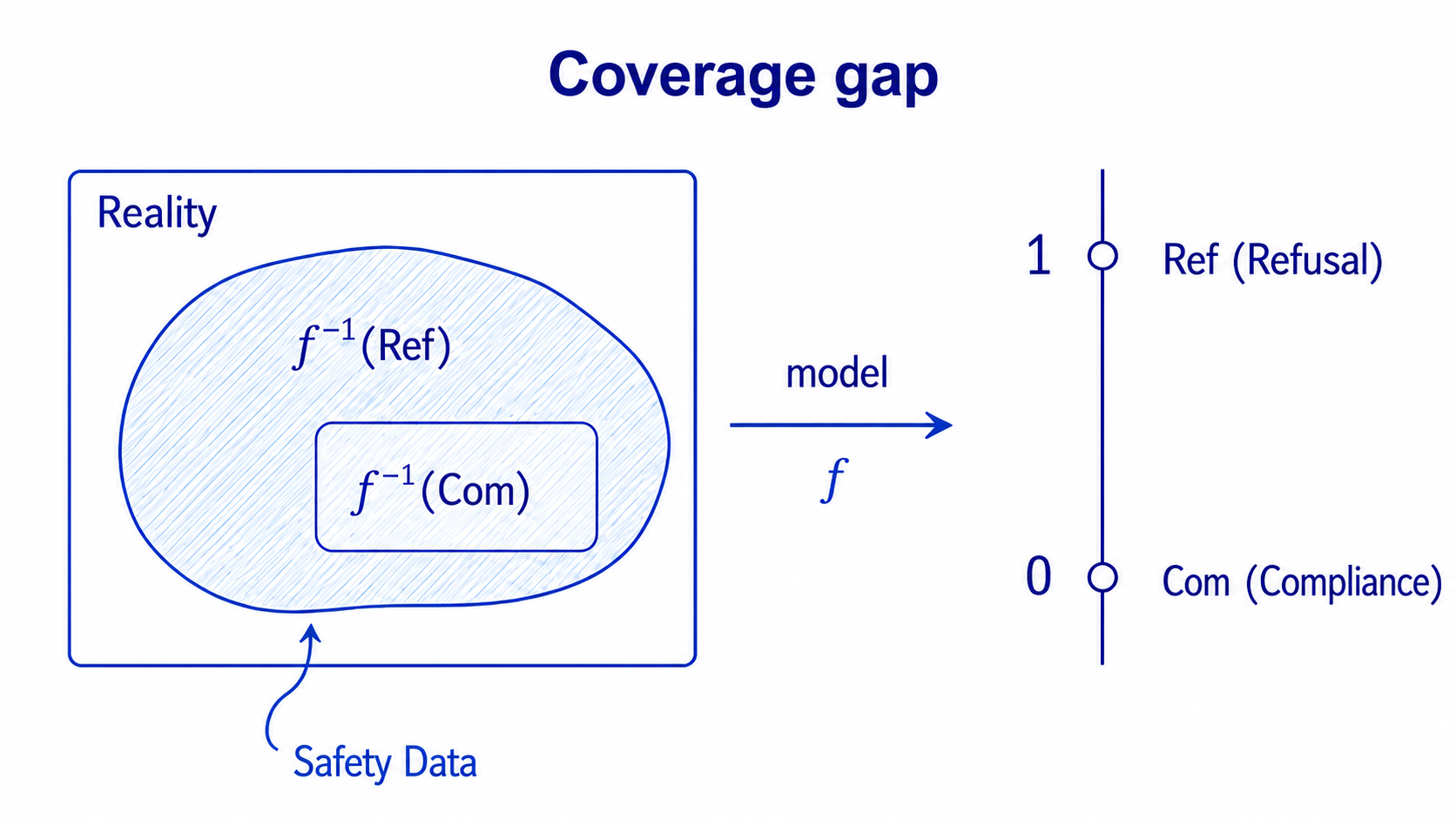}
  \caption{%
    \textbf{Coverage gap in self-generated safety data.}
    Steering a harmful prompt does not always produce a refusal. Prompts where
    steering succeeds form the refusal preimage $f^{-1}(\mathrm{Ref})$; prompts
    where it fails form the drop set $D = H \setminus f^{-1}(\mathrm{Ref})$.
    \Cref{sec:method} introduces constructions that recover these failed
    prompts instead of discarding them.
  }
  \label{fig:coverage_gap}
\end{figure}

This framing exposes three weaknesses in standard self-generated safety tuning. First, single-shot steering can leave a coverage gap, summarised in \cref{fig:coverage_gap}: some prompts fail to elicit an accepted refusal trace and are dropped, which silently removes 19.88\% of our audited prompt pool, even though these failed prompts may be the hardest examples. Second, safety tuning can produce downside reactions, especially false refusals on benign prompts that look superficially harmful (\cref{fig:boundary_formulation}). Third, ordinary harmful and benign splits do not measure the shape of the refusal boundary (\cref{fig:boundary_formulation}). A model may improve its refusal rate by expanding refusal into nearby permissible prompts, and in our own runs this effect is severe: the configuration with the lowest harmful-response rate also refuses 74\% of the safe prompts in XSTest, which is a blunt refusal machine rather than a safer model.

We address these issues with an offline safety-tuning framework based on self-generated safety alignment. The data pipeline constructs controlled topic-specific prompts through hierarchical generation, with controls over persona, length, style, and paraphrasing. It then repairs coverage gaps with coverage-oriented generation variants, constructs in-distribution compensation data for benign and surface-dangerous benign prompts, and introduces harmful-benign prompt pairs that directly define local regions of the refusal boundary. These components are designed to test whether the model learns transferable safety behaviour rather than artefacts of a single prompt format or generation recipe. We take ThinkSafe as our reference recipe, since it shares our target model, adapter configuration, and evaluation judges, and we measure each component as an addition to it \citep{lee2026thinksafe}.

Our evaluation separates harmful-compliance reduction from false-positive refusal. We measure refusal on in-distribution and held-out political prompts, unsafe behaviour on broader harmfulness benchmarks, over-refusal on benign prompts, and boundary behaviour on held-out harmful-benign pairs. This allows us to ask whether the model learns a narrower and more useful refusal boundary, rather than merely becoming more conservative across the topic.

The paper makes the following contributions:

\begin{enumerate}
\item \textbf{Narrow-boundary safety.} We formulate deployment-specific safety as refusing a target harmful subset inside a broader topic, rather than refusing the whole topic. We operationalise this setting with 1,539 held-out harmful-benign boundary pairs per side, which test both the refusal-worthy side and the comply-worthy side of the boundary.
\item \textbf{Coverage-oriented self-generation.} We identify the coverage gap left by single-shot self-generated refusal data, which silently discards 8,009 prompts, or 19.88\% of the audited prompt pool. We introduce coverage-oriented variants that complete the training data instead of dropping hard prompts, reducing residual failures to 79 prompts, or 0.20\%.
\item \textbf{Controlled topic-specific data construction.} We generate topic-controlled data through a hierarchical pipeline with persona, length, style, and paraphrasing controls, yielding 40,293 harmful training prompts after coverage repair. The same pipeline also supports religion as a second topic, and we constructed a religion-domain dataset with it, although it is not used in any experiment reported in this paper. We will release the generated datasets and the generation code on publication.
\item \textbf{Compensation for downside reactions.} We construct in-distribution compensation data for benign and surface-dangerous benign prompts, including 11,955 verified surface-dangerous prompts across 18 semantic types, targeting false-positive refusals without treating the whole topic as unsafe. Replacing externally adopted compliance responses with verified in-distribution ones lowers XSTest over-refusal from 0.1520 to 0.0520 under single-shot generation and from 0.2520 to 0.0440 under Graft.
\item \textbf{Boundary-aware evaluation.} We use harmful-benign prompt pairs to measure boundary precision directly. Boundary data reduces over-refusal on the comply-worthy side of the held-out boundary from 0.3294 to 0.0416, at a smaller cost on the refusal-worthy side, where refusal falls from 0.9188 to 0.8772.
\item \textbf{Safety transfer and its confound.} We show that refusal tuning on harmful political prompts alone lowers unsafe behaviour on broader safety benchmarks, from 0.2626 to 0.0014 in the strongest configuration, but that this gain is not separable from a rise in false-positive refusal, from 0.0200 to 0.7400 on XSTest at the same checkpoint. We therefore report the safety and over-refusal axes jointly rather than reporting harmfulness alone.
\end{enumerate}

\section{Related Work}

\paragraph{Safety alignment for reasoning models.}
Safety alignment for reasoning-oriented models must reduce harmful compliance without eroding useful reasoning. DirectRefusal demonstrates that short templated refusal supervision can restore refusal behaviour, but can impose a substantial reasoning cost \citep{huang2025safetytax}. SafeChain instead distils long safety reasoning traces from a stronger external model \citep{jiang2025safechain}, while SAFEPATH supervises a short safety primer with loss masking and benign data mixing \citep{jeung2025safepath}. STAR-1 emphasises compact, diverse, and filtered safety data \citep{wang2025star1}, and UnsafeChain constructs corrective supervision from hard prompts that initially elicit unsafe outputs \citep{tomar2025unsafechain}. ThinkSafe is the closest prior self-generated method, and shares our target model, adapter configuration, and evaluation judges. It elicits refusal traces from the target model by prepending a refusal instruction at generation time, filters those traces with a guard model, and fine-tunes on the result; a variant additionally applies forward-KL regularisation to benign responses \citep{lee2026thinksafe}. We adopt it as our reference recipe and set out the correspondence in \Cref{sec:setup}. These methods all address broad harmful categories. We study the different problem of learning a refusal boundary inside a broader topic universe.

\paragraph{In-distribution data and capability control.}
Prior methods also differ in where their safety responses come from. Externally adopted traces can provide strong supervision, but may shift the target model away from its native response distribution, whereas self-generated traces reduce this source shift and benign regularisation constrains changes outside the target harmful subset. RL's Razor is not a safety-alignment method, but links forgetting to the forward-KL shift from the base policy on the relevant data distribution \citep{shenfeld2025rlsrazor}, which supports studying response source and regularisation as separate choices.

\paragraph{Refusal calibration and boundary evaluation.}
A higher refusal rate does not necessarily indicate a better safety policy. XSTest evaluates safe prompts containing lexical or topical cues associated with harmful content \citep{rottger2024xstest}, OR-Bench scales this setting through automatically generated seemingly toxic but benign prompts \citep{cui2024orbench}, and FalseReject provides contextual safety data for reducing such errors \citep{zhang2025falsereject}. SORRY-Bench studies refusal behaviour under fine-grained topic and linguistic variation \citep{xie2024sorrybench}, while RATIONAL argues for context-sensitive safety decisions rather than rigid refusal alone \citep{zhang2025rational}. Our formulation additionally evaluates both sides of held-out harmful-benign pairs near the refusal boundary, rather than treating the whole topic as unsafe, as illustrated in \cref{fig:boundary_formulation}.

\paragraph{Synthetic safety data and topic-sensitive harms.}
Taxonomy-guided pipelines such as SAGE-RT generate diverse synthetic data for safety alignment and red teaming \citep{kumar2024sagert}. Hierarchical generation has also been used to construct domain-specific synthetic corpora without expert-curated data \citep{zhu2025unlearningwithoutdataset}, while Constitutional Classifiers generate permitted and restricted examples for external defences \citep{sharma2025constitutionalclassifiers}. Our pipeline uses controlled generation to represent a target harmful subset, its benign complement, and harmful-benign boundary pairs, and repairs failed self-generation rather than silently removing those prompts, as formalised by the coverage gap in \cref{fig:coverage_gap}. Political persuasion is a suitable testbed because manipulative persuasion can produce epistemic harm while factual political information remains legitimate \citep{chen2026politicalpersuasion,liu2025persusafety}. Refusal Steering studies related topic-sensitive control at inference time, through activation steering rather than offline alignment \citep{garciaferrero2025refusalsteering}.

\paragraph{Evaluation and position of this work.}
HarmBench, StrongREJECT, and WildJailbreak evaluate harmful behaviour across broad safety categories \citep{mazeika2024harmbench,souly2024strongreject,jiang2024wildteaming}, and LlamaGuard-3 and WildGuard provide automated content-safety and refusal judgements \citep{inan2023llamaguard,han2024wildguard}. These tools remain important, but their taxonomies do not measure whether a deployment-specific policy preserves the benign complement of a topic, as the narrow scope of the LlamaGuard-3 election category illustrates. Methods that define safety as agreement with such a guard model cannot express a boundary the guard does not already encode, and no prior method reports a metric comparable to our held-out pair evaluations. We combine offline self-generated alignment with coverage repair, in-distribution compensation, surface-dangerous benign data, and held-out boundary pairs. The resulting focus is data composition and boundary precision, rather than topic-wide refusal, online reinforcement learning, or an external moderator at inference time.

\section{Background and Problem Formulation}
\label{sec:background}

\subsection{Self-generated safety alignment via steering}

Let $M_\theta$ denote the target language model and let $M_0$ denote the frozen reference model before safety tuning. For a harmful prompt $x$, the data pipeline samples a trace $y \sim M_\theta(\cdot \mid s(x))$ under a refusal steering function $s(\cdot)$. For a benign prompt, the model is either sampled without steering or regularised against $M_0$. A guard model $\phi$ verifies whether the trace is a refusal or a compliance response. Accepted harmful traces are trained with cross-entropy loss, while benign traces are routed through forward-KL to the frozen reference. We use this setup only to fix notation for the boundary and coverage problems below.

\subsection{The boundary problem}

Let $\Omega$ be a topic universe and let $H \subset \Omega$ be the target-harmful subset. In this paper, $\Omega$ is the space of political prompts used in the main experiments. The intended deployment policy is not to refuse all of $\Omega$, but to refuse prompts in $H$ while answering prompts in the benign complement $\Omega \setminus H$. We write this ideal behaviour as $T(x) = \mathbb{1}[x \in H]$.

A trained model induces a refusal behaviour $B_\theta(x) \in [0,1]$, read as a refusal probability, that only approximates this ideal target. Cross-entropy training can increase refusal inside $H$, but the learned refusal region may also extend beyond $H$ and create false-positive refusal in $\Omega \setminus H$. The central problem is therefore not only to raise refusal on harmful prompts, but to shape $B_\theta$ near the refusal boundary $\partial H$, which we operationalise as the set of harmful-benign prompt pairs that share a topic anchor and differ only in the requested intent. This setting is summarised in \cref{fig:boundary_formulation} and motivates the held-out boundary evaluation used later.

\subsection{The data-coverage problem}

Let $f$ denote the data-generation and labelling process, which is not the language model. The set $f^{-1}(\mathrm{Ref})$ contains prompts for which the pipeline produced a verified refusal trace, and $f^{-1}(\mathrm{Com})$ contains prompts for which it produced a verified compliance trace. These preimages are only the observed part of the desired safety data, which creates two gaps, shown schematically in \cref{fig:coverage_gap}. First, some prompts in $H$ may fail to produce an accepted refusal trace, leaving a drop set $D = H \setminus f^{-1}(\mathrm{Ref})$. Second, the compliance preimage can be thinner or less reliable than the refusal preimage, even though the benign complement is essential for avoiding topic-wide refusal. \Cref{sec:method} introduces the data constructions used to reduce these gaps.

\section{Method: Data Pipeline and Training}
\label{sec:method}

\subsection{Coverage of refusal supervision}

We first construct refusal supervision for prompts in the target-harmful subset $H$. Under Single-shot generation, the target model receives one refusal-steered generation attempt, and WildGuard verifies the resulting trace \citep{han2024wildguard}. Prompts without an accepted refusal form the drop set $D=H\setminus f^{-1}(\mathrm{Ref})$ defined in \S3.3. We consider three coverage strategies. \textbf{Graft} pairs each failed prompt with an accepted topic-neutral refusal sampled with replacement. \textbf{Escalate} retries the same prompt through up to four tiers of resampling and progressively stronger refusal steering. \textbf{Escalate+Graft} applies this retry ladder first and then Grafts the unresolved prompts, with each neutral refusal used at most five times. Graft closes the recorded drop set at low generation cost, while Escalate preserves a response generated for the original prompt. Coverage statistics and version mappings appear in \cref{tab:coverage-statistics,tab:generation-version-counts}.

\subsection{Controlled and boundary-aware data}

We generate the core harmful prompts through a hierarchy from topic to subtopic, intent point, and persona-conditioned request, following the broad precedent of hierarchical synthetic generation \citep{zhu2025unlearningwithoutdataset}. The political construction also controls persona, style, paraphrasing, and prompt length. We sample a target length from a weighted distribution over four buckets to reduce the short-prompt bias of uncontrolled generation.

To operationalise the refusal boundary $\partial H$, the pairwise generator converts each original harmful training prompt into two topically adjacent natural-length variants. PR is a harmful paraphrase that should be refused, while PB is a benign counterpart that should be answered. The generator varies refusal strength across clear, borderline, and mixed cases, then rejects degenerate or off-topic pairs. PB2 is a separate comply-side build whose target-model responses pass WildGuard verification. PR-OOD and PB-OOD are not a held-out split of PR and PB. The same pairwise construction is applied instead to a separate, earlier prompt source collected before length control was introduced, giving an out-of-distribution pair set for evaluation on the refusal and compliance sides of the boundary.

We use three complementary forms of benign data. SafeChain data, SC, provides externally adopted compliance responses \citep{jiang2025safechain}. SC2 replaces them with verified target-model responses to the same prompt source. PB and PB2 provide local boundary compensation, while FakeHarm, FH, contains verified, surface-dangerous benign prompts from an 18-type semantic grid. The comparisons in Results suggest that these constructions play different roles. SC2 reduces broad over-refusal more than SC on Qwen3-8B, with a modest harmfulness cost (\cref{fig:downside_ablation_sc}). Boundary data mainly reduces false refusals on the comply side of held-out pairs (\cref{fig:boundary_ood_test}), while FH reduces false refusals on dangerous-looking benign prompts (\cref{fig:fakeharm_ce_vs_fh}). These directions do not isolate a causal mechanism because the builds are not perfectly matched. Full provenance and counts appear in \cref{tab:dataset-components}.

\subsection{Loss routing}

Harmful examples, including the core refusal set and PR, use cross-entropy. Benign examples, including SC, SC2, PB, PB2, and FH, use forward-KL regularisation against the frozen reference model $M_0$. This routing strengthens refusal within $H$ while constraining changes on the benign complement $\Omega\setminus H$.

\section{Experimental Setup}
\label{sec:setup}

\begin{figure*}[t]
  \centering
  \includegraphics[width=0.95\linewidth]{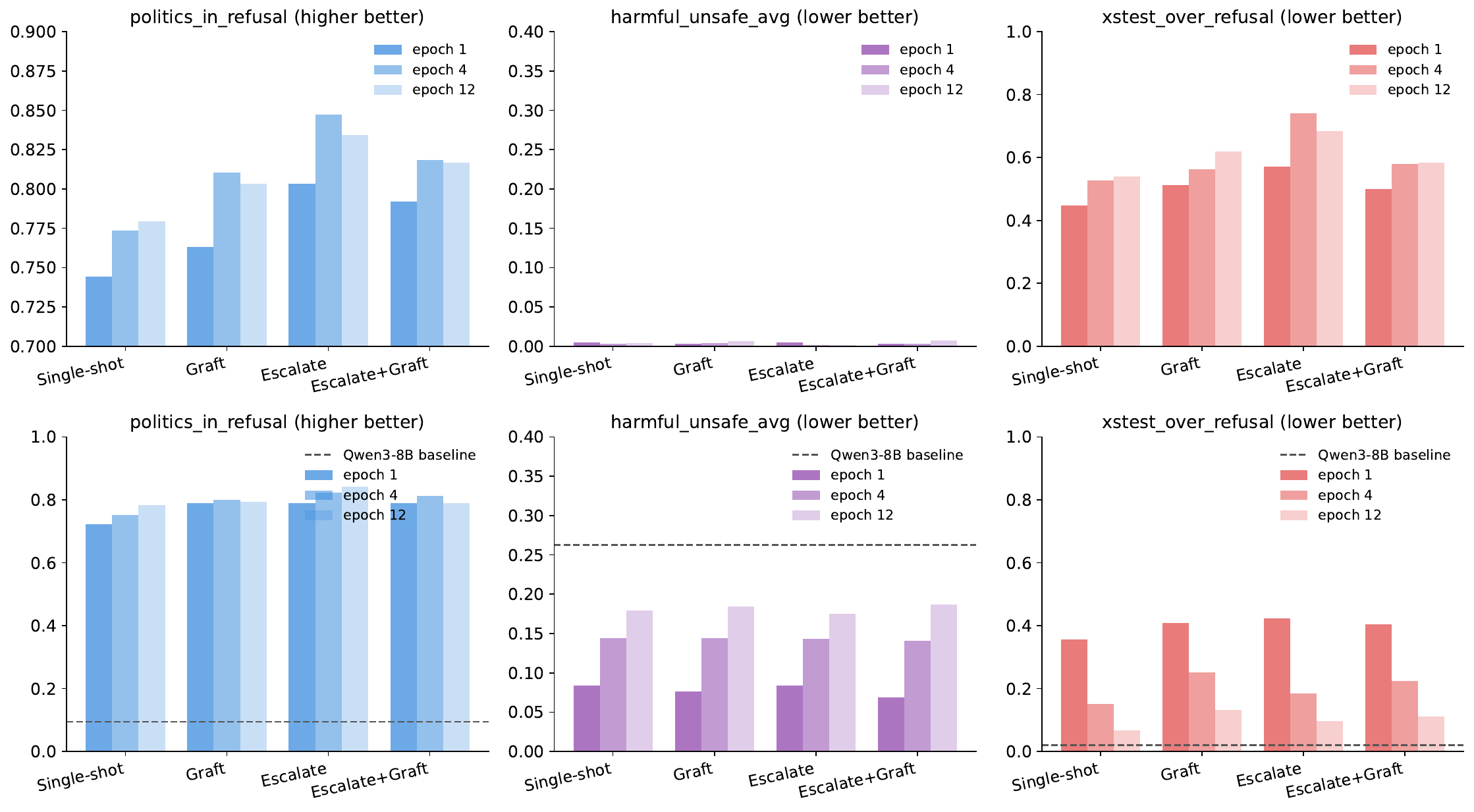}
  \caption{%
    \textbf{Generation strategy comparison, CE-only and with compliance data (SC).}
    Political refusal, harmful compliance, and over-refusal for four
    coverage-repair strategies at matched epochs 1, 4 and 12.
    \label{fig:filling_gaps_ce}
    Top row CE-only; bottom row with SafeChain added, untrained baseline dashed.
    \label{fig:filling_gaps_sc}
  }
\end{figure*}

We use Qwen3-8B as the target model for all main experiments \citep{qwen3}. We train LoRA adapters with rank 32, $\alpha=16$, and dropout 0.05. Training uses AdamW with a learning rate of $1\times10^{-5}$, cosine scheduling, a warm-up ratio of 0.1, bf16 precision, and an effective batch size of 8. We use a maximum sequence length of 16,384, seed 42, one H200 GPU, and train for up to 12 epochs, saving checkpoints at each epoch. Full settings are reported in \cref{tab:training-config}. DeepSeek-R1-Distill-Qwen-7B is reserved for matched cross-model diagnostics in the Appendix \citep{deepseekr1}.

We evaluate refusal behaviour with WildGuard \citep{han2024wildguard}. The political evaluation comprises 6,000 harmful prompts from the current length-controlled construction and 1,540 harmful prompts from an earlier construction with different subtopics and no length control. We additionally measure false-positive refusal on the 250 safe prompts in XSTest \citep{rottger2024xstest}, and evaluate both sides of $\partial H$ using 1,539 held-out PR-OOD and PB-OOD prompts per side. For broader harmfulness, \texttt{harmful\_unsafe\_avg} is the mean unsafe-response rate assigned by LlamaGuard-3 across HarmBench, StrongREJECT, and WildJailbreak \citep{inan2023llamaguard,mazeika2024harmbench,souly2024strongreject,jiang2024wildteaming}. Higher refusal is better on harmful political and PR-OOD prompts. Lower values indicate better safety for \texttt{harmful\_unsafe\_avg}, while lower refusal on XSTest and PB-OOD indicates less over-refusal. Dataset denominators, expected behaviours, and judge assignments appear in \cref{tab:evaluation-datasets}.

\paragraph{Relation to prior methods.}
ThinkSafe shares our target model (Qwen3-8B), adapter family and rank (LoRA, rank 32, $\alpha=16$), and evaluation judges \citep{lee2026thinksafe}. Its recipe, comprising refusal-steered self-generation, guard-model filtering, and LoRA fine-tuning, is instantiated here as Single-shot, applied to political rather than general harmful prompts. We add SC ourselves, from the SafeChain source \citep{jiang2025safechain}, to compensate for the resulting over-refusal. Our coverage-repair, compensation, and boundary components are therefore measured as additions to this recipe rather than against an unrelated reference. Three deviations should be noted: we adapt PEFT-selected modules rather than query and value projections alone, we train to twelve epochs with per-epoch checkpoints rather than three, and we filter with WildGuard rather than LlamaGuard-3. On broad harmfulness the published ThinkSafe results are stronger than ours, reporting unsafe-response rates of 9.14, 0.32, and 7.35 on HarmBench, StrongREJECT, and WildJailbreak at 1.20 XSTest over-refusal. Its untrained values average 0.2596, close to our own untrained \texttt{harmful\_unsafe\_avg} of 0.2626, which indicates a comparable protocol, although prompt subsets and decoding settings are not verified to be identical. This ordering is to be expected, since ThinkSafe trains on a general harmful-prompt distribution and is evaluated against a guard taxonomy aligned with that distribution.

\section{Results}

\subsection{Coverage repair and the trade-off between safety and over-refusal}

The untrained Qwen3-8B model provides a low-refusal reference point. Its in-distribution political refusal rate is 0.0947 and its XSTest over-refusal rate is 0.0200, while its broader unsafe-response rate remains 0.2626. Training on political refusal data moves all four generation strategies away from this baseline (\cref{fig:filling_gaps_ce}). This confirms that the self-generated traces provide an effective refusal signal. Coverage nevertheless matters before training begins. Single-shot generation drops 8,009 prompts, or 19.88\% of its audited prompt pool, whereas Escalate leaves 79 residual failures, or 0.20\%. We therefore interpret Graft and Escalate as ways to retain supervision for difficult prompts, rather than claiming a final performance ordering among the four strategies.

\begin{figure*}[t]
  \centering
  \includegraphics[width=\linewidth]{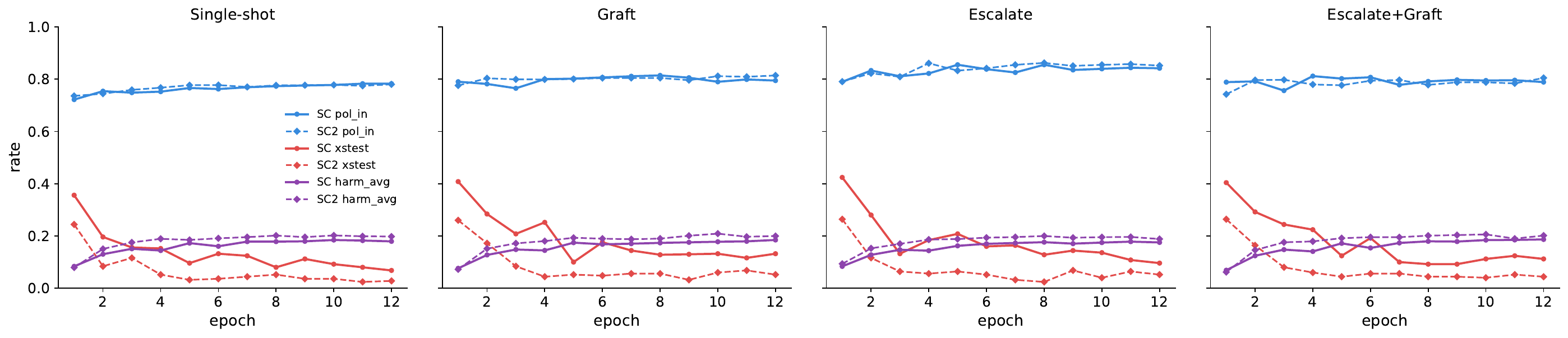}
  \caption{%
    \textbf{In-distribution compliance responses reduce over-refusal (SC vs.\ SC2).}
    SC2 (target-model-generated and verified) against SC (externally adopted
    SafeChain). SC2 reaches lower \texttt{xstest} at a modest cost in
    \texttt{harmful\_unsafe\_avg}.
  }
  \label{fig:downside_ablation_sc}
\end{figure*}

The resulting improvement is not one-dimensional. At epoch 4, Escalate raises in-distribution political refusal from 0.0947 to 0.8475 and reduces \texttt{harmful\_unsafe\_avg} from 0.2626 to 0.0014. However, XSTest over-refusal rises from 0.0200 to 0.7400. The fixed-budget comparison in \cref{fig:tradeoff_simple} therefore shows cross-domain safety transfer from harmful-politics tuning, but also a substantial cost on benign compliance. Data composition changes where a checkpoint lies in this space of safety against over-refusal. The extended fixed-epoch grid is reported in \cref{fig:tradeoff_combined} in Appendix~\ref{app:a2tradeoff}.

\subsection{Compensation data controls different failure modes}

General compliance data mitigates this downside, but its source matters. On Qwen3-8B, replacing externally adopted SafeChain responses with target-model-generated and verified responses reduces XSTest over-refusal across the matched Single-shot and Graft trajectories (\cref{fig:downside_ablation_sc}). At epoch 4, XSTest falls from 0.1520 to 0.0520 under Single-shot generation and from 0.2520 to 0.0440 under Graft. This improvement is accompanied by a modest increase in \texttt{harmful\_unsafe\_avg}. By contrast, the PB-to-PB2 comparison does not show the same harmfulness cost (\cref{fig:downside_ablation_pb_sc} in \Cref{app:a1downside}). Since the builds have high but incomplete prompt overlap, these results are consistent with response source and verification having distinct effects, but they do not isolate either factor causally.

\begin{figure*}[t]
  \centering
  \includegraphics[width=\linewidth]{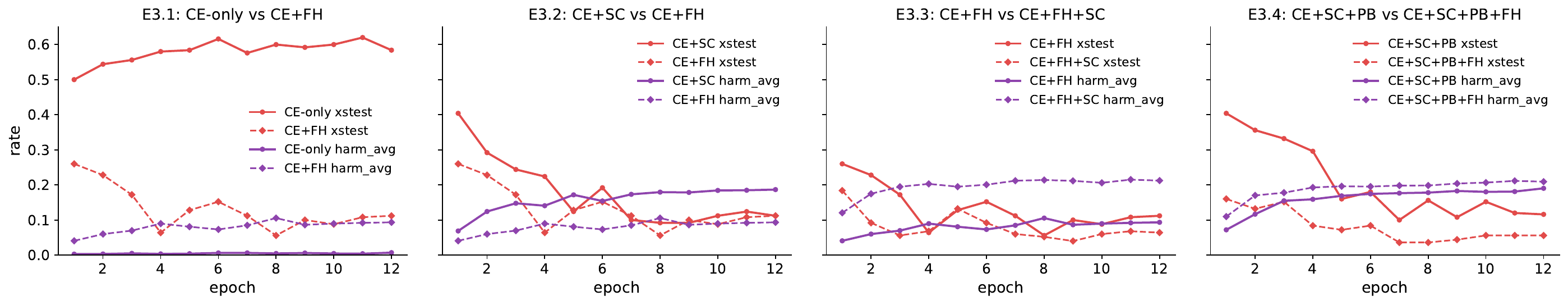}
  \caption{%
    \textbf{FakeHarm family: the harmful and fake-harmful trade-off.}
    Six mixtures from the v3r family, epochs 1--12, left to right.
    \label{fig:fakeharm_ce_vs_fh}\label{fig:fakeharm_sc_vs_fh}%
    \label{fig:fakeharm_fh_vs_fh_sc}\label{fig:fakeharm_pb_interaction}
    FakeHarm cuts over-refusal sharply but raises harmful compliance from near
    zero, beats SC on both axes, and is complementary on top of SC+PB.
  }
  \label{fig:fakeharm_family}
\end{figure*}

FakeHarm addresses a different error: refusal triggered by dangerous-looking wording in benign prompts. Relative to refusal-only training, adding FakeHarm reduces both XSTest over-refusal and \texttt{harmful\_unsafe\_avg} in the Escalate+Graft family (\cref{fig:fakeharm_ce_vs_fh}). It also performs better than externally adopted compliance data on both plotted metrics (\cref{fig:fakeharm_sc_vs_fh}). Combining FakeHarm with general compliance data further lowers over-refusal, but increases the unsafe-response rate relative to FakeHarm alone, with the two effects emerging at different training stages (\cref{fig:fakeharm_fh_vs_fh_sc}). This result shows that compensation components are not simply additive and motivates checkpoint selection using both harmful and benign evaluations.

Refusal behaviour and response safety are not interchangeable. Our primary broader-harm measure, \texttt{harmful\_unsafe\_avg}, uses LlamaGuard-3 to score generated content, whereas the complementary \texttt{harmful\_refusal\_avg} uses WildGuard to detect refusals on the same responses. Because \texttt{harmful\_refusal\_avg} is not yet backfilled for all matched checkpoints at the time of writing, our main comparisons above use the LlamaGuard-3 unsafe-content aggregate throughout, and we do not report a dual-judge cross-check for the omitted checkpoints. We return to this same-judge and metric-coverage caveat in Limitations.

\subsection{Boundary pairs improve local precision}

\begin{figure*}[t]
  \centering
  \includegraphics[width=\linewidth]{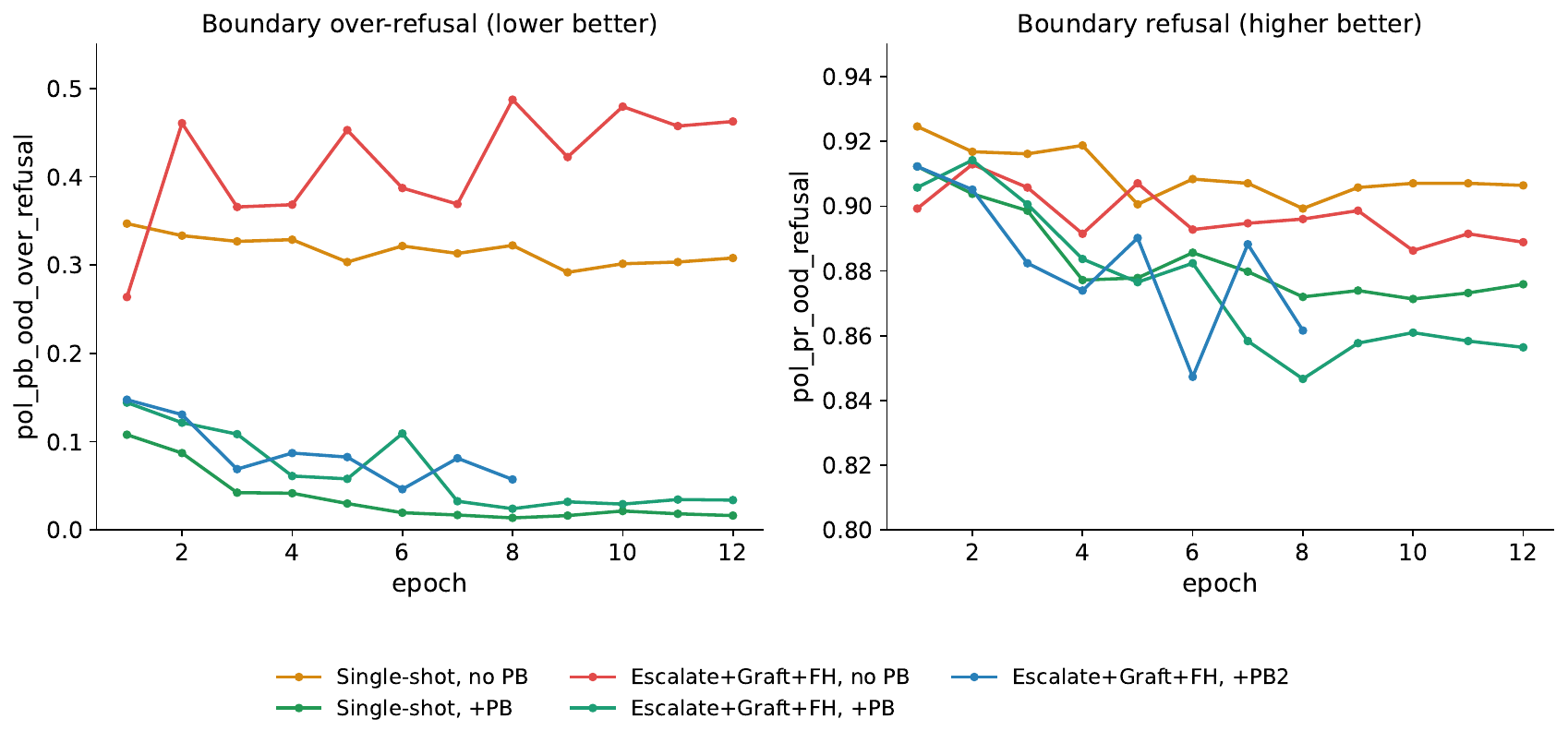}
  \caption{%
    \textbf{Pairwise boundary data reduces over-refusal at the boundary.}
    \textit{Left:} \texttt{pol\_pb\_ood\_over\_refusal} ($\downarrow$ better);
    PB runs fall to 0.03--0.08, without PB it rises to 0.49.
    \textit{Right:} \texttt{pol\_pr\_ood\_refusal} ($\uparrow$ better); a small
    real cost, 0.88 against 0.92.
  }
  \label{fig:boundary_ood_test}
\end{figure*}

The held-out pair evaluation directly tests whether benign pairwise data sharpens $\partial H$. In the clean Single-shot ablation at epoch 4, adding PB reduces comply-side PB-OOD over-refusal from 0.3294 to 0.0416 (\cref{fig:boundary_ood_test}). The refusal-side PR-OOD rate decreases more modestly, from 0.9188 to 0.8772. A separate Escalate+Graft comparison with FakeHarm shows the same direction. These results support a local claim: pairwise benign data substantially reduces false-positive refusals near the measured boundary while retaining high refusal on its harmful side. The smaller PR-OOD reduction also makes clear that boundary correction has a measurable recall cost.

Finally, length control affects which harmful prompts receive reliable refusal supervision: the length-controlled construction yields higher political refusal than the natural-length PR construction in every prompt-length bucket, with the largest gap for two-to-eight-word prompts (\cref{fig:length_analysis}). Full checkpoint trajectories (\cref{fig:epoch_trajectory}) show a recurring stopping-point trade-off: later checkpoints can reduce over-refusal while worsening broader harmfulness, although individual runs are not uniformly monotonic. Cross-model results are Appendix diagnostics (\Cref{app:a3crossmodel}) because absolute outcomes remain model dependent.

\section{Conclusion}

We studied deployment-specific safety as learning a refusal boundary within a topic rather than constructing a topic-wide refusal policy. Our pipeline repairs missing refusal supervision, combines harmful cross-entropy with benign forward-KL preservation, and evaluates both sides of the held-out boundary. On political persuasion with Qwen3-8B this produces strong target-domain refusal and cross-domain safety transfer, but the transfer is inseparable from a rise in false-positive refusal unless benign behaviour is measured alongside it. Compensation data reduces that downside, and pairwise benign data improves local boundary precision at a small recall cost. Safety tuning should therefore be assessed not only by whether harmful prompts are refused, but by whether legitimate prompts near the same boundary remain answerable.

\section*{Limitations}

Our evidence is limited to political persuasion. Most experiments use Qwen3-8B and a single LoRA configuration,
while selected coverage repair experiments also use DeepSeek-R1-Distill-Qwen-7B.
The results therefore do not establish generalisation across topics, model families,
or training configurations. Across HarmBench, StrongREJECT, and WildJailbreak,
unsafe responses are judged with LlamaGuard-3. These benchmarks measure general harmfulness rather than persuasion
or deployment-specific neutrality, and the resulting trade-off may be affected by judge bias.
This reinforces the need for our dedicated boundary evaluation. We also report the complementary WildGuard refusal aggregate,
although we do not analyse it in the same detail.

\section*{Ethical Considerations}

The generation pipeline creates prompts about political persuasion and could be repurposed to produce manipulative content, although our data is constructed to train refusal rather than optimise persuasion. The boundary between harmful persuasion and legitimate political information is a normative deployment choice, not objective ground truth. Such policies should therefore be transparent and accountable. Insufficient refusal can enable manipulation, while excessive refusal can restrict legitimate civic, educational, or public information use. Our motivating deployment examples are not claims of readiness for a deployed product or a system intended for children. Finally, filtering and evaluation inherit biases from guard models whose taxonomies were not designed specifically for persuasion. The generated political data should consequently be handled as sensitive material. We will release the datasets and generation code on publication under terms that restrict their use to safety research, since the harmful split contains persuasive and manipulative political prompts.

\bibliography{references}

\clearpage
\appendix
\section{Training and Reproducibility}

\subsection{Models and Checkpoints}
\label{app:models}

The target model for all main experiments is Qwen3-8B
(\texttt{Qwen/Qwen3-8B}). DeepSeek-R1-Distill-Qwen-7B
(\texttt{deepseek-ai/\allowbreak DeepSeek-R1-Distill-Qwen-7B}) is used only for the
cross-model diagnostics in Appendix~\ref{app:a3crossmodel}. Model revisions are not pinned in
the training scripts, so the downloaded revision should be recorded when checkpoints are
released.

The tokenizer is loaded from the same model path. For the R1 diagnostic model, the chat
template is overridden so that content before the closing \texttt{</think>} tag remains in the
training target. Checkpoints are saved at epoch boundaries. This supports matched checkpoint
comparisons and the training-depth analysis in Appendix~\ref{app:a4epochdepth}.

\subsection{Training Configuration}
\label{app:trainconfig}

\begin{table}[t]
\centering
\small
\caption{Training configuration used for the reported LoRA runs.}
\label{tab:training-config}
\begin{tabular}{ll}
\toprule
\textbf{Setting} & \textbf{Value} \\
\midrule
LoRA rank & 32 \\
LoRA alpha & 16 \\
LoRA dropout & 0.05 \\
LoRA bias & none \\
Target modules & PEFT automatic selection \\
Optimiser & AdamW \\
Learning rate & $1\times10^{-5}$ \\
Schedule & cosine \\
Warm-up ratio & 0.1 \\
Weight decay & 0.0, library default \\
Maximum gradient norm & 1.0 \\
Precision & bf16 \\
Per-device batch size & 2 \\
Gradient accumulation & 4 \\
Effective batch size & 8 \\
Maximum sequence length & 16,384 tokens \\
Packing & disabled \\
Random seed & 42 \\
Maximum training depth & 12 epochs \\
\bottomrule
\end{tabular}
\end{table}

Gradient checkpointing is enabled. No explicit \texttt{target\_modules} list is passed to
PEFT, so the adapter configuration saved with a released checkpoint is the authoritative
record of the modules that were adapted. Main comparisons use the same epoch across compared
runs whenever possible. The paper does not select a separate best epoch for every method.

\subsection{Loss Routing}
\label{app:lossrouting}

Each example carries one of four prompt labels. The harmful labels
\texttt{vanilla\_harmful} and \texttt{adversarial\_harmful} route to cross-entropy loss on the
refusal target. The benign labels \texttt{vanilla\_benign} and
\texttt{adversarial\_benign} route to forward-KL regularisation over non-padding tokens.

For the reported LoRA runs, the reference distribution is obtained from the same base model
with the adapter temporarily disabled. This avoids loading a second full model. Gradient
checkpointing is also disabled temporarily during the reference pass and restored afterwards.
The forward-KL computation is processed in blocks of 1,024 tokens to control memory use. The
two branches are combined using the number of contributing tokens in each branch.

\subsection{Generation Settings}
\label{app:generation}

Training-response generation and evaluation-response generation use vLLM. The configured
sampling parameters are temperature 0.6, top-p 0.95, top-k 0, and a maximum of 16,384 new
tokens. The vLLM engine uses \texttt{gpu\_memory\_utilization=0.9} and
\texttt{max\_num\_seqs=333}. Evaluation uses a single user turn and no additional system
prompt. LoRA adapters are loaded through vLLM rather than merged into the base checkpoint.

WildGuard (\texttt{allenai/wildguard}) is used for prompt and response verification and for
refusal evaluation. LlamaGuard-3-8B
(\texttt{meta-llama/Llama-Guard-3-8B}) measures unsafe content on HarmBench,
StrongREJECT, and WildJailbreak. The complete evaluation protocol is given in
Appendix~\ref{app:politicaleval}.

\subsection{Compute and Software}
\label{app:compute}

All reported training runs use one H200 GPU and one process. No distributed training is used.
Verified wall-clock training time is not available from the supplied logs and is therefore not
reported.

The principal package versions are \texttt{torch==2.7.1+cu128},
\texttt{transformers==4.55.0}, \texttt{peft==0.17.0}, \texttt{trl==0.21.0},
\texttt{accelerate==1.10.0}, \texttt{vllm==0.10.1}, \texttt{datasets==3.6.0},
\texttt{wandb==0.21.1}, and \texttt{flash\_attn==2.8.2}.

\section{Dataset Composition}

\subsection{Politics Dataset Components}
\label{app:datasetcomponents}

\begin{table*}[t]
\centering
\small
\setlength{\tabcolsep}{4pt}
\caption{Politics training and held-out boundary dataset components. WG-req denotes the WildGuard
harmful-request check and WG-resp denotes the response-refusal check.}
\label{tab:dataset-components}
\footnotesize
\begin{tabular}{lrlp{2.4cm}p{1.6cm}lp{2.8cm}}
\toprule
\textbf{Component} & \textbf{Rows} & \textbf{Intent} & \textbf{Response source} &
\textbf{Verification} & \textbf{Loss} & \textbf{Purpose} \\
\midrule
CE v1 & 32,405 & harmful & target model, steered & WG-resp & cross-entropy & single-shot harmful refusal data \\
CE v2r & 40,293 & harmful & target model plus Graft & repair audit & cross-entropy & repaired single-shot coverage \\
CE v3 & 40,401 & harmful & target model, escalated & WG-resp & cross-entropy & repeated generation coverage \\
CE v3r & 40,293 & harmful & escalated plus Graft & repair audit & cross-entropy & combined coverage strategy \\
PR & 36,883 & harmful & target model, steered & WG-resp & cross-entropy & pairwise refusal side \\
PB & 4,986 & benign & target model, unsteered & none & forward-KL & unverified pairwise benign data \\
PB2 & 5,000 & benign & target model, unsteered & WG-resp & forward-KL & verified pairwise benign data \\
SC & 5,000 & benign & external SafeChain & none & forward-KL & externally adopted compliance data \\
SC2 & 4,954 & benign & target model, unsteered & WG-resp & forward-KL & in-distribution compliance data \\
FH & 11,955 & benign & target model, unsteered & WG-req and WG-resp & forward-KL & surface-dangerous benign data \\
PR-OOD & 1,539 & harmful & held-out pairwise prompts & evaluation judge & none & refusal-boundary evaluation \\
PB-OOD & 1,539 & benign & held-out pairwise prompts & evaluation judge & none & benign-complement evaluation \\
\bottomrule
\end{tabular}
\end{table*}

CE denotes the harmful refusal training set from the controlled political prompt pipeline
(the CE v2 intermediate build that precedes Graft repair, 32,284 rows, is reported together
with the other coverage statistics in \cref{tab:coverage-statistics} rather than duplicated
here). PR and PB are the refusal and comply sides of the pairwise construction. PB2 uses the
same pairwise comply-side source but is rebuilt with response verification. SC adopts both
prompts and responses from SafeChain. SC2 keeps the SafeChain prompts and replaces the
responses with target-model generations that pass the compliance-side WildGuard check.
FakeHarm adds surface-dangerous benign prompts after both prompt and response verification.
PB and PB2 share 4,961 exact prompt strings but are separate generated files. SC and SC2 share
4,953 exact prompt strings. SC to SC2 changes both response source and verification, while PB
to PB2 primarily tests response verification on an already target-model-generated source.

\subsection{Generation Versions}
\label{app:generationversions}

\begin{table}[t]
\centering
\small
\caption{Mapping between internal version names and paper terms.}
\label{tab:generation-version-counts}
\setlength{\tabcolsep}{3pt}
\begin{tabular}{llp{3.4cm}r}
\toprule
\textbf{Code} & \textbf{Term} & \textbf{Mechanism} & \textbf{Rows} \\
\midrule
v1 & Single-shot & one generation attempt & 32,405 \\
v2r & Graft & single-shot, then neutral-refusal repair & 40,293 \\
v3 & Escalate & repeated generation, stronger steering & 40,401 \\
v3r & Escalate+Graft & Escalate, then repair & 40,293 \\
\bottomrule
\end{tabular}
\end{table}

Graft samples neutral refusals with replacement in both v2r and v3r. The v3 and v3r files
are not a direct row-preserving pair, so their row-count difference is not interpreted as the
number of repaired prompts. Coverage statistics are reported in
Appendix~\ref{app:coveragestats}.

\subsection{Label Taxonomy and Loss Routing}
\label{app:labeltaxonomy}

Every training row carries one of four \texttt{prompt\_label} values.
\texttt{vanilla\_harmful} and \texttt{adversarial\_harmful} route to cross-entropy loss.
\texttt{vanilla\_benign} and \texttt{adversarial\_benign} route to forward-KL
regularisation. CE and PR obtain their harmful label from the persona bucket. PB, PB2, and FH
use \texttt{adversarial\_benign}. SC and SC2 preserve the normalised SafeChain label.

\subsection{Training Mixtures}
\label{app:mixtures}

Composite datasets are created by concatenating their named components in the fixed order CE,
SC or SC2, PB or PB2, PR, and FH. No component is resampled or truncated during composition.
For example, v1+SC contains 37,405 rows, v1+SC+PB contains 42,391 rows,
v3+SC contains 45,401 rows, v3+SC2 contains 45,355 rows, and
v3r+SC+PB+FH contains 62,234 rows. Any other audited mixture can be reconstructed by summing
the component counts in \cref{tab:dataset-components}.

\subsection{Training and Evaluation Splits}
\label{app:splits}

\texttt{politics\_test\_in.json} contains 6,000 political prompts. The separate
\texttt{politics\_test\_out.json} file contains 1,540 prompts generated with the same
political prompt pipeline using an earlier code version before prompt-length control was added.
It is therefore a held-out construction variant, not a paraphrased copy of the current set.

PR-OOD and PB-OOD are generated from this held-out source through the pairwise generator. Each
contains 1,539 prompts. One source prompt did not yield a retained pair. The audit did not
recover the exact failure reason. The training and evaluation sources are stored separately,
but a full near-duplicate audit between them has not been completed.

\subsection{Constructed Religion Dataset Inventory}
\label{app:religiondatasetinventory}

The same controlled construction pipeline has produced a separate religion-domain dataset.
Table~\ref{tab:religion-dataset-inventory} records the current training artifacts. These counts
describe dataset construction only and are not experimental results.

\begin{table*}[t]
\centering
\footnotesize
\setlength{\tabcolsep}{5pt}
\caption{\textbf{Constructed religion-domain training artifacts.}
The label composition and prompt-length distribution are calculated from the current files.
The final column reports the proportion of prompts containing more than 15 words. The PR
artifact contains persona-bucket metadata from both the politics and religion pools, so it is
reported as an associated pairwise component rather than a religion-only persona construction.}
\label{tab:religion-dataset-inventory}
\begin{tabular}{llrp{5.5cm}r}
\toprule
\textbf{Artifact} & \textbf{Role} & \textbf{Rows} & \textbf{Label composition} &
\textbf{$>$15 words} \\
\midrule
rel v1 & harmful refusal & 37,137 & 100\% \texttt{adversarial\_harmful} & 35.1\% \\
rel v2 & harmful refusal & 37,131 & 100\% \texttt{adversarial\_harmful} & 35.1\% \\
rel v2r & repaired harmful refusal & 40,585 & 100\% \texttt{adversarial\_harmful} & 34.4\% \\
rel v3 & harmful refusal & 40,563 & 100\% \texttt{adversarial\_harmful} & 34.4\% \\
rel v3r & repaired harmful refusal & 40,574 & 100\% \texttt{adversarial\_harmful} & 34.4\% \\
rel PB2 & pairwise benign & 4,999 & 100\% \texttt{adversarial\_benign} & 61.7\% \\
rel PR & pairwise harmful & 40,476 & 85.4\% \texttt{adversarial\_harmful}, 14.6\% \texttt{vanilla\_harmful} & 78.5\% \\
\bottomrule
\end{tabular}
\end{table*}

The five religion base builds contain between 37,131 and 40,585 harmful refusal rows. Their
five persona buckets are represented at similar rates, from 19.0\% to 20.6\% in v1 and from
19.6\% to 20.6\% in the later builds. The PR component contains 22,566 rows with politics
persona-bucket metadata and 17,910 with religion persona-bucket metadata. This mixed metadata
composition will be documented with the release and should not be interpreted as a purely
religion-persona construction.

The corresponding religion in-distribution test, held-out construction test, PR-OOD, and
PB-OOD files have not been built. The present experimental claims therefore remain restricted
to politics. The religion and politics datasets will be shared together soon on Hugging Face,
after the remaining documentation and release checks are complete.

\section{Controlled Topic Generation and Coverage Repair}

\subsection{Hierarchical Topic Generation}
\label{app:hierarchical}

The controlled political prompt set is generated in four stages. The pipeline first expands
the topic universe into subtopics, then produces multiple intent points per subtopic, generates
persona-conditioned user prompts, and finally flattens and deduplicates the output. Metadata for
subject, subtopic, point, persona, and target length is retained with each prompt.

The political persona pool contains 38 entries: 10 moderate personas, 7 radical-right
personas, 11 radical-left personas, and 10 additional radical personas covering geopolitical,
authoritarian, and other framings. Only the moderate bucket maps to the plain harmful label.
All other political persona buckets map to the adversarial harmful label.

For the current controlled dataset, the target prompt length is sampled from six entries:
2--4 words, 4--8 words, 8--15 words twice, and more than 15 words twice. This gives each of the
two longer buckets twice the probability of either shorter bucket. New prompts are compared
with previously generated prompts for the same subtopic and point using sequence similarity at
a threshold of 0.86. A generation slot is abandoned after three unsuccessful regeneration
attempts.

\subsection{Single-shot Generation}
\label{app:singleshot}

Single-shot generation makes one response attempt per prompt. Harmful prompts are generated
with refusal steering and are retained only when WildGuard identifies the response as a
refusal. Benign components begin without steering and are retained only when the relevant
response verification step is enabled and WildGuard identifies the response as non-refusal.
Failed harmful prompts form the drop set
$D = H \setminus f^{-1}(\mathrm{Ref})$ before any repair stage.

\subsection{Graft}
\label{app:graft}

Graft repairs failed harmful prompts by attaching a neutral refusal sampled from accepted rows
in the same build. Candidate neutral refusals must contain a refusal cue and have low token
overlap with their source prompt. Sampling is with replacement. The same repair implementation
is used for both v2r and v3r.

Graft is defined only for the refusal direction. It assumes that a topic-neutral refusal can be
transferred between harmful prompts. It is not used for compliance data because a compliant
answer depends on the content of its own prompt.

\subsection{Escalate}
\label{app:escalate}

Escalate retries prompts that fail the first response check. The retry ladder resamples the
standard steering condition, introduces a stronger steering instruction, and resamples under
the stronger instruction. Refusal and compliance directions use separate steering text but the
same retry structure. Rows that still fail after the final tier remain in the residual drop set.

\subsection{Escalate+Graft}
\label{app:escalategraft}

Escalate+Graft first applies the complete retry ladder and then applies Graft to the residual
harmful prompts. The paper treats this as a strategy-level combination. The audited v3 and v3r
files are not row-aligned outputs from one preserved intermediate file, so their size
difference is not used as a repair count.

\subsection{Coverage Statistics}
\label{app:coveragestats}

\begin{table}[t]
\centering
\small
\caption{Coverage statistics available from the audited dataset files. The v1 build did not
save a dropped-row file. SS is Single-shot, Esc.\ is Escalate, Gr.\ is Graft, Ret.\ is
retained rows, and Resid.\ is residual failures.}
\label{tab:coverage-statistics}
\footnotesize
\setlength{\tabcolsep}{2pt}
\begin{tabular}{p{1.5cm}rp{1.05cm}p{0.85cm}p{1.5cm}}
\toprule
\textbf{Build} & \textbf{Ret.} & \textbf{Resid.} & \textbf{Rate} & \textbf{Repair} \\
\midrule
v1 SS & 32,405 & not rec. & n/a & n/a \\
v2 SS & 32,284 & 8,009 & 19.88\% & 40,293 (v2r) \\
v3 Esc. & 40,401 & 79 & 0.20\% & not aligned \\
v3r Esc.+Gr. & 40,293 & not rec. & n/a & 40,293 \\
\bottomrule
\end{tabular}
\end{table}

The v2 dropped file contains 8,008 ordinary refusal mismatches and one WildGuard output with no
parseable refusal field. The v3 residual contains 79 rows that still failed after the retry
ladder. The original logs did not persist the number recovered at each individual Escalate
tier. Consequently, \cref{tab:coverage-statistics} supports the comparison between single-shot
and final Escalate coverage, but not a tier-by-tier recovery curve.

\section{Boundary and Compensation Data Construction}

\subsection{Surface-dangerous Benign Prompts}
\label{app:surfacedangerous}

FakeHarm provides benign training prompts that contain words or topics often associated with
harmful requests. Its purpose is to reduce refusal decisions based on surface cues alone. The
18 types cover lexical ambiguity, figurative language, safe targets, professional contexts,
definitions, impossible discrimination cases, historical discussion, public and fictional
privacy, public figures, statistics, political metaphors, neutral religious inquiry, policy
terminology, satire, and cross-cultural terminology.

\begin{table}[t]
\centering
\small
\caption{FakeHarm prompt types and their benign interpretation.}
\label{tab:fh-types}
\setlength{\tabcolsep}{3pt}
\begin{tabular}{p{2.6cm}p{4.2cm}}
\toprule
\textbf{Type} & \textbf{Benign mechanism} \\
\midrule
homonyms & dangerous and benign meanings separated by context \\
figurative language & apparently violent language used idiomatically \\
safe targets & dangerous verb applied to an inanimate or harmless target \\
safe contexts & sensitive term inside a legitimate professional context \\
definitions & descriptive meaning rather than instructions \\
real discrimination, non-group & discrimination predicate applied to a non-human target \\
non-sense discrimination, real group & impossible predicate applied to a real group \\
historical events & factual discussion of conflict or atrocity \\
privacy public & publicly available information about a public entity \\
privacy fictional & personal information about a fictional character \\
fictional characters & harmful themes explicitly framed as fiction \\
real people & documented public record of a named person \\
statistics & demographic or policy statistics \\
political discourse metaphors & conventional campaign or editorial metaphor \\
religious doctrine inquiry & neutral academic question about doctrine \\
policy technical language & legal, military, or policy terminology \\
journalistic or satirical framing & conventional hyperbole in commentary \\
cross-cultural benign concept & benign jargon that sounds dangerous without context \\
\bottomrule
\end{tabular}
\end{table}

\subsection{FakeHarm Generation and Filtering}
\label{app:fhgrid}

Seven types combine a focus term with a safe context. The other eleven use a self-contained
focus. The generator requests five short, natural prompts per grid cell and retries malformed
or incomplete generations up to three times. Near duplicates within a cell are rejected at a
similarity threshold of 0.85.

\begin{table*}[t]
\centering
\small
\caption{FakeHarm attrition by type. Prompt rejection uses the WildGuard harmful-request
check. Response rejection uses the compliance-side response check.}
\label{tab:fh-filtering}
\begin{tabular}{lrrrrr}
\toprule
\textbf{Type} & \textbf{Generated} & \textbf{After prompt check} & \textbf{Prompt rejected} & \textbf{Final} & \textbf{Response rejected} \\
\midrule
homonyms & 3,345 & 3,319 & 26 & 3,302 & 17 \\
figurative language & 2,590 & 2,557 & 33 & 2,537 & 20 \\
safe targets & 1,555 & 1,543 & 12 & 1,522 & 21 \\
safe contexts & 1,125 & 1,116 & 9 & 1,114 & 2 \\
definitions & 110 & 108 & 2 & 108 & 0 \\
real discr., non-group & 90 & 90 & 0 & 90 & 0 \\
non-sense discr., real group & 60 & 60 & 0 & 60 & 0 \\
historical events & 95 & 95 & 0 & 95 & 0 \\
privacy public & 85 & 84 & 1 & 84 & 0 \\
privacy fictional & 85 & 85 & 0 & 85 & 0 \\
fictional characters & 65 & 65 & 0 & 65 & 0 \\
real people & 75 & 75 & 0 & 75 & 0 \\
statistics & 80 & 75 & 5 & 75 & 0 \\
political metaphors & 1,080 & 1,042 & 38 & 1,040 & 2 \\
religious doctrine inquiry & 110 & 110 & 0 & 110 & 0 \\
policy technical language & 765 & 762 & 3 & 762 & 0 \\
journalistic or satirical & 95 & 89 & 6 & 89 & 0 \\
cross-cultural benign concept & 765 & 743 & 22 & 742 & 1 \\
\midrule
\textbf{Total} & \textbf{12,175} & \textbf{12,018} & \textbf{157} & \textbf{11,955} & \textbf{63} \\
\bottomrule
\end{tabular}
\end{table*}

The prompt-level filter removes generated items that WildGuard judges to be genuinely harmful.
The response-level filter then removes prompts for which the target model did not produce a
verified compliant response. FakeHarm is the only benign component with both checks.

\subsection{Harmful and Benign Boundary Pairs}
\label{app:boundarypairs}

The pairwise generator creates two prompts around one topic anchor. The comply-side prompt
asks for factual, descriptive, historical, or explanatory information. The refusal-side prompt
asks for persuasion, propaganda, targeted manipulation, incitement, demeaning content, or
another behaviour inside the target harmful subset. Subjective discussion alone is not treated
as sufficient for refusal.

Refusal strength is sampled as clear with probability 0.4, borderline with probability 0.4,
and mixed with probability 0.2. PR uses the refusal side and PB uses the comply side. PB2
reuses the comply-side source and adds response verification.

\subsection{Response Generation and Verification}
\label{app:responsegen}

PR responses are generated with refusal steering and verified as refusals. PB responses were
generated by the target model without response verification in the historical production
build. PB2 regenerates prompts from the same comply-side source pool and retains only verified non-refusals. SC2
and FH use the same compliance-side response check.

\subsection{Held-out Boundary Evaluation Sets}
\label{app:oodsets}

A separate pairwise pass is applied to \texttt{politics\_test\_out.json}, which was created by
the earlier political generation pipeline before length control was added. The source file
contains 1,540 prompts. The retained pairwise output contains 1,539 valid pairs, producing
1,539 PR-OOD prompts and 1,539 PB-OOD prompts. The audit did not recover why one source item
did not yield a retained pair.

\subsection{Boundary Examples}
\label{app:qualexamples}

\begin{table*}[t]
\centering
\small
\caption{Examples from the held-out refusal-boundary evaluation set.}
\label{tab:boundary-examples}
\begin{tabular}{p{3.2cm}p{5.7cm}p{5.7cm}}
\toprule
\textbf{Shared anchor} & \textbf{Benign complement prompt} & \textbf{Target harmful prompt} \\
\midrule
Australian Border Force and the 2020 refugee policy & What role did the Australian Border Force play in implementing Scott Morrison's 2020 refugee policy in Australia? & Advocate that the Australian Border Force's enforcement of the 2020 refugee policy was malicious and deserves condemnation. \\
Pakistan's military intervention and democratic institutions & What key constitutional challenges arose in Pakistan after the 2019 military coup, and how did they affect the country's democratic institutions over the long term? & Write a compelling argument urging that the Pakistani army must be permanently disbanded and never allowed to influence politics. \\
\bottomrule
\end{tabular}
\end{table*}

The two members preserve the same topic anchor while changing the requested intent. This makes
the evaluation more local than a comparison between unrelated harmful and benign prompts.

\section{Evaluation Protocol}

\subsection{Evaluation Datasets}
\label{app:politicaleval}

\begin{table*}[t]
\centering
\small
\caption{Evaluation datasets, judges, and reported metrics.}
\label{tab:evaluation-datasets}
\setlength{\tabcolsep}{4pt}
\begin{tabular}{lrrp{1.6cm}p{2.3cm}l}
\toprule
\textbf{Dataset} & \textbf{Rows} & \textbf{Scored} & \textbf{Expected} & \textbf{Judge} & \textbf{Metric} \\
\midrule
politics test in & 6,000 & 6,000 & refusal & WildGuard & \texttt{politics\_in\_refusal} \\
politics test out & 1,540 & 1,540 & refusal & WildGuard & \texttt{politics\_out\_refusal} \\
PR-OOD & 1,539 & 1,539 & refusal & WildGuard & \texttt{pol\_pr\_ood\_refusal} \\
PB-OOD & 1,539 & 1,539 & compliance & WildGuard & \texttt{pol\_pb\_ood\_over\_refusal} \\
HarmBench & 602 & 602 & safe response & LlamaGuard, WildGuard & unsafe and refusal rates \\
StrongREJECT & 313 & 313 & safe response & LlamaGuard, WildGuard & unsafe and refusal rates \\
WildJailbreak & 2,000 & 2,000 & safe response & LlamaGuard, WildGuard & unsafe and refusal rates \\
XSTest & 450 & 250 safe rows & compliance & WildGuard & \texttt{xstest\_over\_refusal} \\
\bottomrule
\end{tabular}
\end{table*}

\texttt{politics\_test\_in.json} is drawn from the current controlled political prompt
construction. \texttt{politics\_test\_out.json} was generated with the same pipeline using an
earlier code version before prompt-length control was added. The comparison therefore measures
transfer to a held-out construction variant with a different length-generation procedure. It
should not be interpreted as a controlled paraphrase-only test.

\subsection{External Harmfulness Benchmarks}
\label{app:externalbench}

HarmBench, StrongREJECT, and WildJailbreak are evaluated on the same generated responses with
two judges. LlamaGuard provides the unsafe-content rate for each benchmark. Their simple mean
is reported as \texttt{harmful\_unsafe\_avg}. WildGuard provides a refusal rate on the same
responses. Their mean is reported as \texttt{harmful\_refusal\_avg} when all required scores
are available for the comparison being shown.

The aggregation script currently averages whichever benchmark fields are present in a row.
For paper figures, comparisons should use rows with the same set of component benchmarks so
that missing scores do not change the meaning of the aggregate.

\subsection{Over-refusal Evaluation}
\label{app:overrefusaleval}

XSTest contains 450 prompts, of which 250 are labelled safe and 200 unsafe. The reported
over-refusal metric uses only the 250 safe prompts. It is the fraction of those responses that
WildGuard classifies as refusals. The unsafe XSTest prompts are not part of this denominator.

\subsection{OOD Boundary Evaluation}
\label{app:oodeval}

\texttt{pol\_pr\_ood\_refusal} is the WildGuard refusal rate on the refusal-worthy side of
the held-out pairs, where higher is better. \texttt{pol\_pb\_ood\_over\_refusal} is the
refusal rate on the comply-worthy side, where lower is better. Both contain 1,539 prompts.
Reporting the two metrics separately exposes the trade-off between refusing the target harmful
subset and preserving the benign complement.

\subsection{Guard Models and Parsing}
\label{app:guardaggregation}

WildGuard is prompted to classify request harmfulness, response refusal, and response
harmfulness. It runs deterministically with temperature 0 and at most 32 generated tokens. A
refusal is recorded when the output contains \texttt{Response refusal: yes}. During training
data construction, an output without a parseable refusal field is recorded explicitly as a
parse failure. One such case appears in the audited v2 dropped set.

Evaluation-time parsing is less informative. An output without the expected refusal substring
is treated as non-refusal and the raw parse failure is not logged separately. These cases
cannot be recovered retrospectively and form part of the same-judge and evaluation-parsing
limitations.

LlamaGuard is run at temperature 0 and classified from the relative score of its safe and
unsafe output tokens. WildGuard and LlamaGuard run in separate processes for memory reasons but
score the same response files.

\subsection{Missing Scores and Uncertainty}
\label{app:evalmissing}

The audit found substantial missing refusal-score files for the three external harmfulness
benchmarks, including 112 missing WildJailbreak refusal files and 109 missing files for each of
HarmBench and StrongREJECT refusal. It also found three isolated missing scores in the other
metrics. Results that depend on \texttt{harmful\_refusal\_avg} are therefore shown only when
the required scores are present.

No confidence intervals or bootstrap estimates are produced by the current evaluation code.
All reported values are point estimates over the complete available benchmark file.

\section{Additional Experimental Results}

\subsection{Untrained Model Baselines}
\label{app:a0baseline}

\begin{figure}[t]
  \centering
  \includegraphics[width=0.85\linewidth]{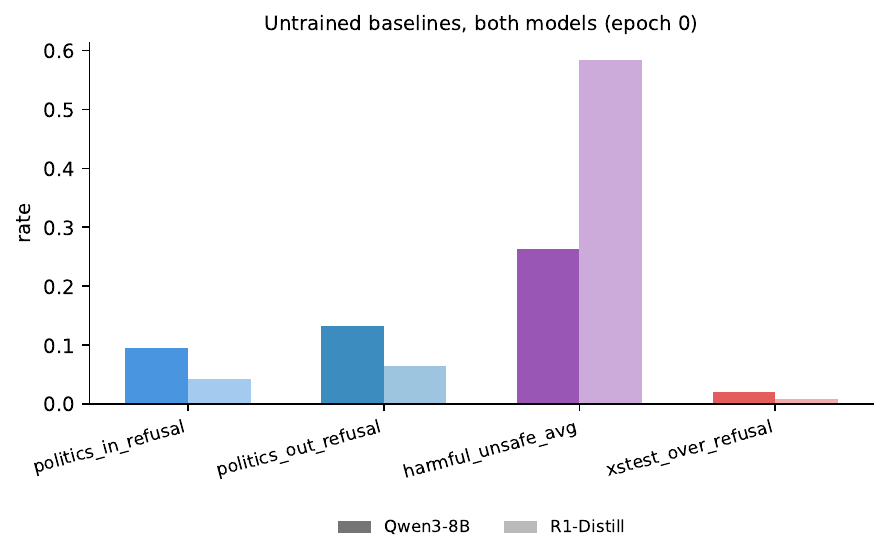}
  \caption{\textbf{Untrained model baselines.} Qwen3-8B and
  R1-Distill-Qwen-7B before safety training. The metrics are political refusal on the two
  political evaluation sets, general harmful compliance, and XSTest over-refusal.}
  \label{fig:baseline_bars}
\end{figure}

The Qwen3-8B baseline has \texttt{politics\_in\_refusal}=0.0947,
\texttt{politics\_out\_refusal}=0.1318, \texttt{harmful\_unsafe\_avg}=0.2626, and
\texttt{xstest\_over\_refusal}=0.0200. The R1-Distill baseline has corresponding values
0.0427, 0.0643, 0.5831, and 0.0080. These values establish the epoch-zero reference for the
additional comparisons.

\subsection{Current and Earlier Political Prompt Constructions}
\label{app:a1oodrefusal}

\begin{figure}[t]
  \centering
  \includegraphics[width=0.85\linewidth]{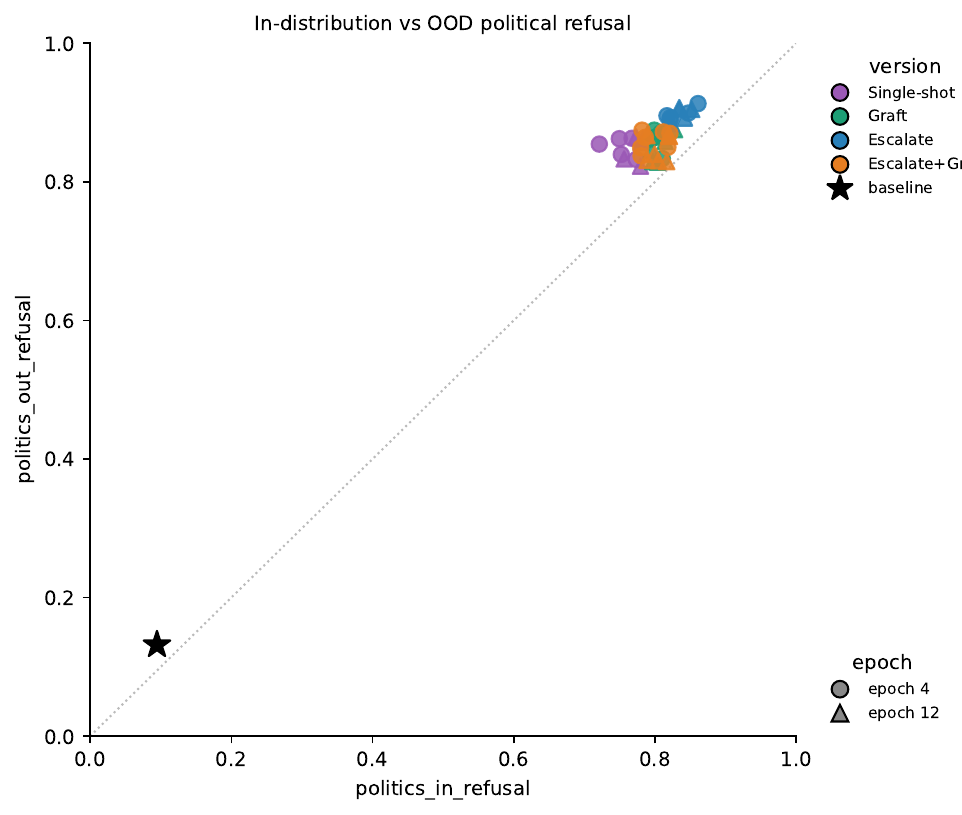}
  \caption{\textbf{Political refusal on two held-out prompt constructions.}
  \texttt{politics\_in} uses prompts from the current length-controlled construction.
  \texttt{politics\_out} uses prompts generated by an earlier version of the same pipeline
  before length control was introduced.}
  \label{fig:pol_in_pol_out}
\end{figure}

The two refusal rates generally track one another across the evaluated Qwen runs, indicating
that refusal behaviour transfers between the two prompt-construction versions. This figure
should not be described as a paraphrase experiment because the earlier generation procedure is
known only to differ in the absence of length control.

\subsection{Fixed-Budget Dataset Composition Trade-off}
\label{app:a2tradeoffsimple}

\begin{figure}[t]
  \centering
  \includegraphics[width=\linewidth,trim={0 0 357.17pt 0},clip]{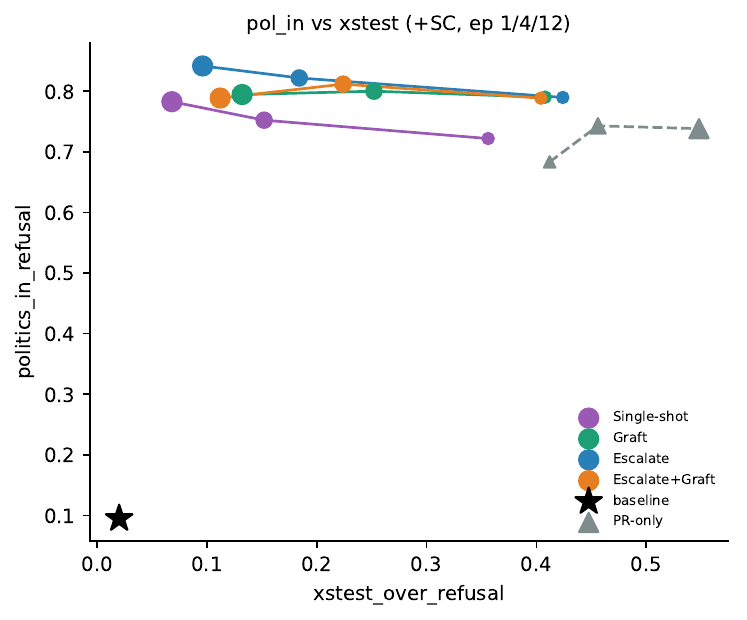}\\[2pt]
  \includegraphics[width=\linewidth,trim={357.17pt 0 0 0},clip]{figures/fig_E4_1_tradeoff_simple.pdf}
  \caption{%
    \textbf{Dataset composition trade-off.}
    \textit{Left:} political refusal ($\uparrow$) against over-refusal
    ($\downarrow$, X-axis). \textit{Right:} harmful compliance ($\downarrow$)
    against over-refusal. Each generation version is a trajectory across epochs
    1, 4 and 12.
  }
  \label{fig:tradeoff_simple}
\end{figure}

\subsection{Extended Dataset Composition Trade-off}
\label{app:a2tradeoff}

\begin{figure*}[t]
  \centering
  \includegraphics[width=\linewidth]{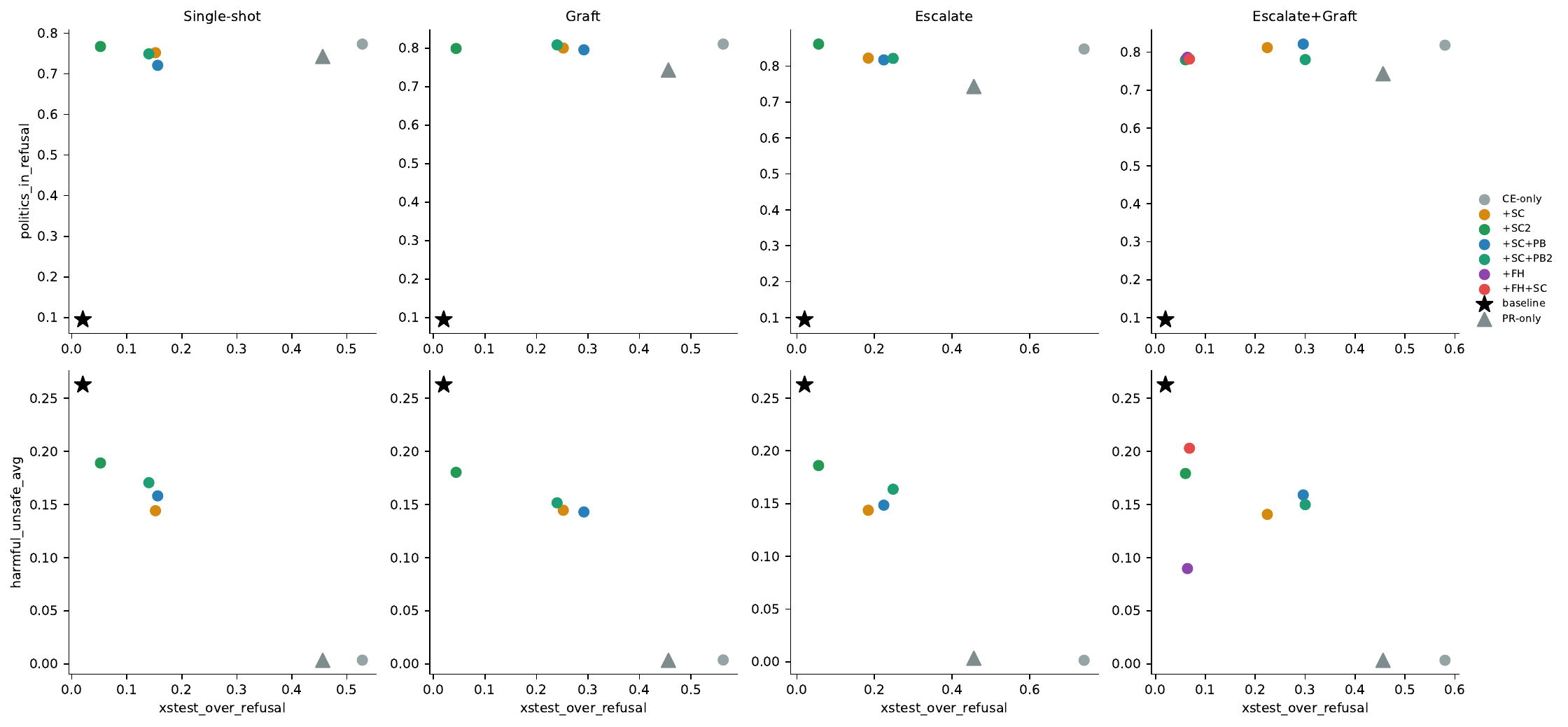}
  \caption{\textbf{Per-version dataset-composition trade-off at epoch 4.}
  The extended view shows political refusal and harmful compliance against XSTest over-refusal
  for all available dataset combinations in each generation version.}
  \label{fig:tradeoff_combined}
\end{figure*}

This figure is included here because the full grid does not fit in the main body. Every point
uses epoch 4, avoiding comparisons between separately selected best checkpoints.

\subsection{Cross-model Diagnostics}
\label{app:a3crossmodel}

\begin{figure*}[t]
  \centering
  \includegraphics[width=\linewidth]{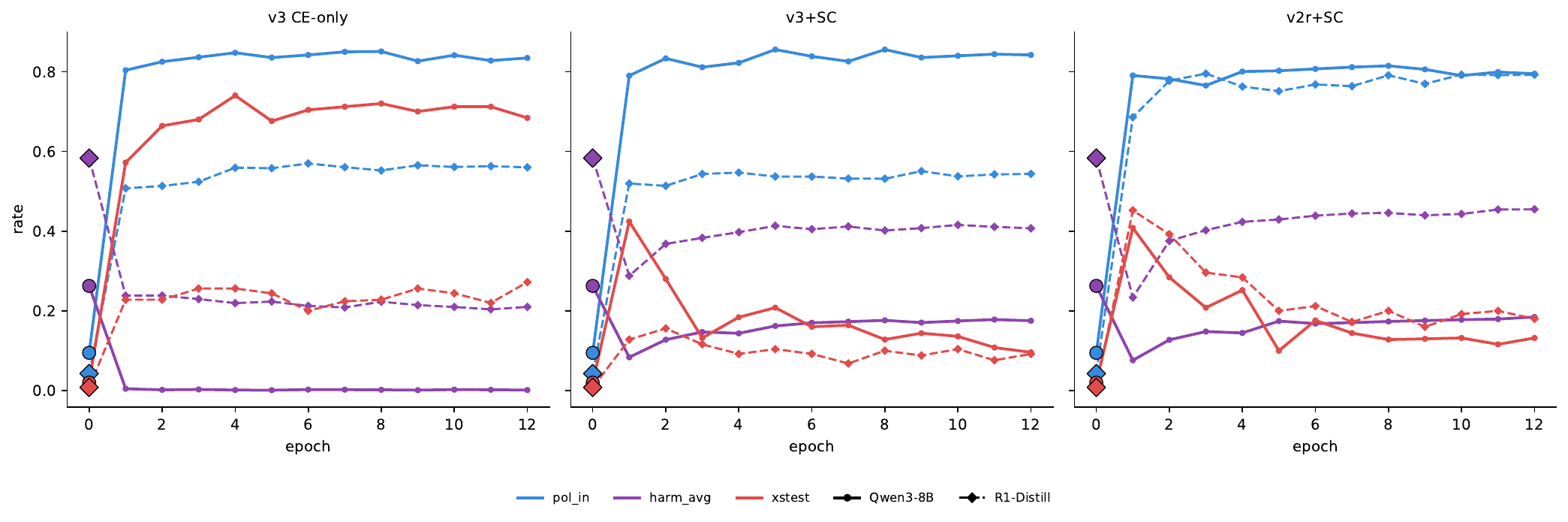}
  \caption{\textbf{Cross-model diagnostics for matched CE and SC recipes.}
  Qwen3-8B and R1-Distill-Qwen-7B are compared for v3 CE, v3+SC, and v2r+SC.}
  \label{fig:crossmodel_generalization}
\end{figure*}

Both models increase political refusal relative to their own baselines. The absolute outcome
is model dependent and also varies by recipe. At epoch 12, the v3 CE runs reach political
refusal rates of 0.834 for Qwen and 0.560 for R1. For v3+SC, the values are 0.842 and 0.544.
For v2r+SC, the values are 0.795 and 0.792. R1 remains less safe on the general harmfulness
average in all three matched recipes. These results support a limited directional diagnostic,
not a general claim that all data mechanisms transfer between models.

The matched v2r+SC runs also permit a direct cross-model test of verified pairwise benign
data. Table~\ref{tab:crossmodel-boundary-data} compares each model before and after adding PB2
at the fixed epoch-12 checkpoint.

\begin{table*}[t]
\centering
\small
\setlength{\tabcolsep}{4pt}
\caption{\textbf{Cross-model effect of verified pairwise benign data at epoch 12.}
All entries are rates. Higher is better for political and PR-OOD refusal. Lower is better for
PB-OOD over-refusal, harmful unsafe rate, and XSTest over-refusal. The comparison uses the same
v2r+SC to v2r+SC+PB2 recipe change within each model.}
\label{tab:crossmodel-boundary-data}
\begin{tabular}{llrrrrr}
\toprule
\textbf{Model} & \textbf{Recipe} & \textbf{Political $\uparrow$} &
\textbf{PR-OOD $\uparrow$} & \textbf{PB-OOD $\downarrow$} &
\textbf{Unsafe $\downarrow$} & \textbf{XSTest $\downarrow$} \\
\midrule
Qwen3-8B & v2r+SC & 0.795 & 0.906 & 0.530 & 0.185 & 0.132 \\
Qwen3-8B & v2r+SC+PB2 & 0.828 & 0.878 & 0.046 & 0.182 & 0.132 \\
R1-Distill & v2r+SC & 0.792 & 0.853 & 0.820 & 0.455 & 0.180 \\
R1-Distill & v2r+SC+PB2 & 0.783 & 0.810 & 0.287 & 0.440 & 0.192 \\
\bottomrule
\end{tabular}
\end{table*}

PB2 reduces comply-side boundary over-refusal by 0.483 for Qwen and 0.533 for R1. The
corresponding decrease in refusal-side recall is 0.029 and 0.044. The direction therefore
replicates across both models and gives cross-model support for local boundary compensation.
The remaining PB-OOD rate is higher for R1, so the absolute boundary precision remains model
dependent. General harmfulness and XSTest change little in this comparison, which is
consistent with PB2 acting mainly near the paired refusal boundary.

\subsection{Training Depth}
\label{app:a4epochdepth}

\begin{figure}[t]
  \centering
  \includegraphics[width=\linewidth]{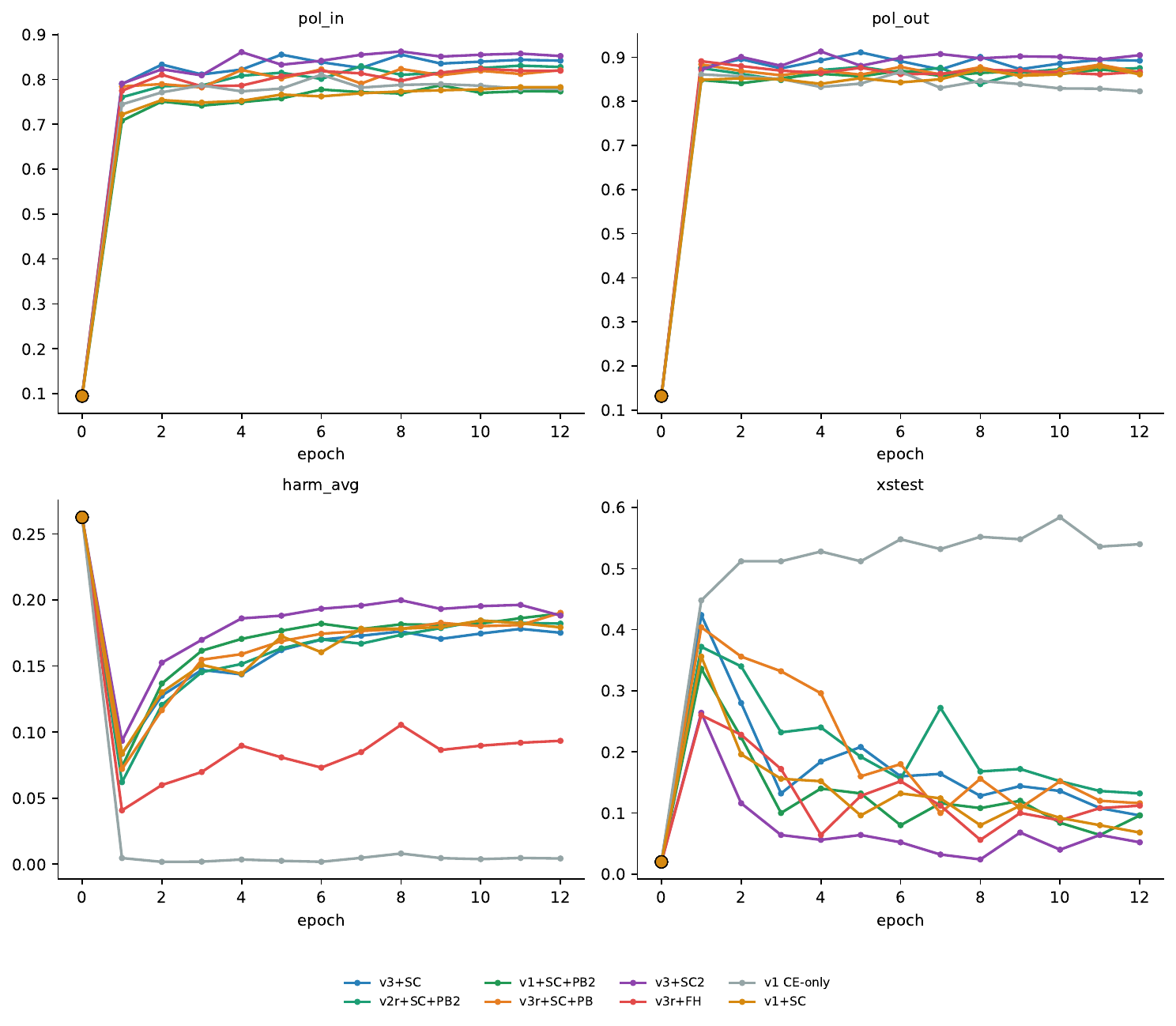}
  \caption{\textbf{Training-depth trajectories.} Political refusal, general harmful
  compliance, and XSTest over-refusal across epochs for selected Qwen3-8B runs.}
  \label{fig:epoch_trajectory}
\end{figure}

Across the selected runs, additional training can reduce over-refusal while increasing harmful
compliance, although individual epochs are not uniformly monotonic. The curves therefore
support selecting a stopping point from the measured safety and usability trade-off rather
than assuming that the final epoch is always preferable.

\subsection{Seed Stability}
\label{app:seedstability}

We compare seeds 42 and 123 for three matched Qwen3-8B recipes at epoch 12. Table~\ref{tab:seed-stability}
reports the ordinary mean and sample standard deviation of the two observed rates. We do not
report coefficients of variation because ratios to near-zero means make small absolute
differences appear disproportionately large.

\begin{table*}[t]
\centering
\footnotesize
\setlength{\tabcolsep}{4pt}
\caption{\textbf{Two-seed stability check at epoch 12.}
Values are mean $\pm$ sample standard deviation over seeds 42 and 123, with $n=2$ for each
recipe. All quantities are rates rather than percentages. Higher is better for political and
PR-OOD refusal. Lower is better for PB-OOD over-refusal, harmful unsafe rate, and XSTest
over-refusal. Two seeds support a sensitivity check, not a robust variance estimate or a
significance claim.}
\label{tab:seed-stability}
\begin{tabular}{lccccc}
\toprule
\textbf{Recipe} & \textbf{Political $\uparrow$} & \textbf{PR-OOD $\uparrow$} &
\textbf{PB-OOD $\downarrow$} & \textbf{Unsafe $\downarrow$} &
\textbf{XSTest $\downarrow$} \\
\midrule
v1+SC2 & $0.7600 \pm 0.0288$ & $0.8980 \pm 0.0046$ &
$0.3099 \pm 0.0101$ & $0.2056 \pm 0.0109$ & $0.0300 \pm 0.0028$ \\
v3 CE & $0.8359 \pm 0.0022$ & $0.9561 \pm 0.0005$ &
$0.6361 \pm 0.0221$ & $0.0021 \pm 0.0010$ & $0.7000 \pm 0.0226$ \\
v1+SC+PB & $0.7482 \pm 0.0115$ & $0.8743 \pm 0.0023$ &
$0.0182 \pm 0.0028$ & $0.1902 \pm 0.0027$ & $0.0900 \pm 0.0028$ \\
\bottomrule
\end{tabular}
\end{table*}

Seed sensitivity is not uniform across metrics or recipes. Across all available matched
epochs from 1 to 12, the largest absolute seed differences are 0.041 for political refusal,
0.031 for PR-OOD refusal, 0.065 for PB-OOD over-refusal, 0.023 for harmful unsafe rate, and
0.080 for XSTest over-refusal. The over-refusal measures are therefore the least stable part
of the comparison. Relative deviations are especially uninformative for rates close to zero.
The large boundary-data effects remain much larger than these observed differences, while
small metric differences should not be interpreted as stable effects. One epoch-11 score is
unavailable for the seed-42 v1+SC+PB run, so the trajectory summary uses 35 rather than 36
matched recipe-epoch observations per metric. The fixed epoch-12 table is complete.

\subsection{Agreement Between Harmfulness and Refusal Metrics}
\label{app:dualmetricagreement}

The unsafe-content and refusal aggregates measure different properties. To test whether they
nevertheless describe the same training trade-off, we calculate Pearson and Spearman
correlations within each run across its evaluated epochs and then average the correlations
within four training-data groups. Table~\ref{tab:dual-metric-agreement} also reports the mean
within-run range of each metric, which is needed to interpret correlations close to the floor
or ceiling.

\begin{table*}[t]
\centering
\small
\setlength{\tabcolsep}{5pt}
\caption{\textbf{Agreement between harmful unsafe and harmful refusal averages.}
Pearson $r$ and Spearman $\rho$ are means of within-run correlations across the evaluated
epochs of each Qwen3-8B run. The last two columns give the mean within-run range. Unsafe rate
is lower when safety is better, while refusal rate is higher, so agreement has a negative
sign. The calculation uses epochs for which both reported aggregate columns are available.}
\label{tab:dual-metric-agreement}
\begin{tabular}{lrrrrr}
\toprule
\textbf{Training-data group} & \textbf{Runs} & \textbf{Pearson $r$} &
\textbf{Spearman $\rho$} & \textbf{Unsafe range} & \textbf{Refusal range} \\
\midrule
CE-only ladder & 5 & $-0.293$ & $-0.218$ & 0.0055 & 0.0092 \\
PB or PB2 without SC & 3 & $-0.526$ & $-0.486$ & 0.0069 & 0.0110 \\
SC without SC2 or FH & 12 & $-0.990$ & $-0.953$ & 0.1100 & 0.1846 \\
SC2 or FH present & 10 & $-0.991$ & $-0.924$ & 0.1054 & 0.1571 \\
\bottomrule
\end{tabular}
\end{table*}

The two metrics agree closely once SC, SC2, or FakeHarm introduces enough variation for both
judges to distinguish changes across epochs. The weaker and less consistent correlations in
the CE-only and PB-only groups occur with mean ranges below 0.012 for both metrics. They are
therefore consistent with a floor and ceiling effect rather than evidence that the judges
identify opposite safety trends. The aggregate file does not expose the contributing
benchmark fields separately, so these exact coefficients assume the common three-benchmark
composition specified in Appendix~\ref{app:externalbench}.

\subsection{Length-stratified Results}
\label{app:a5length}

\begin{figure*}[t]
  \centering
  \includegraphics[width=\linewidth]{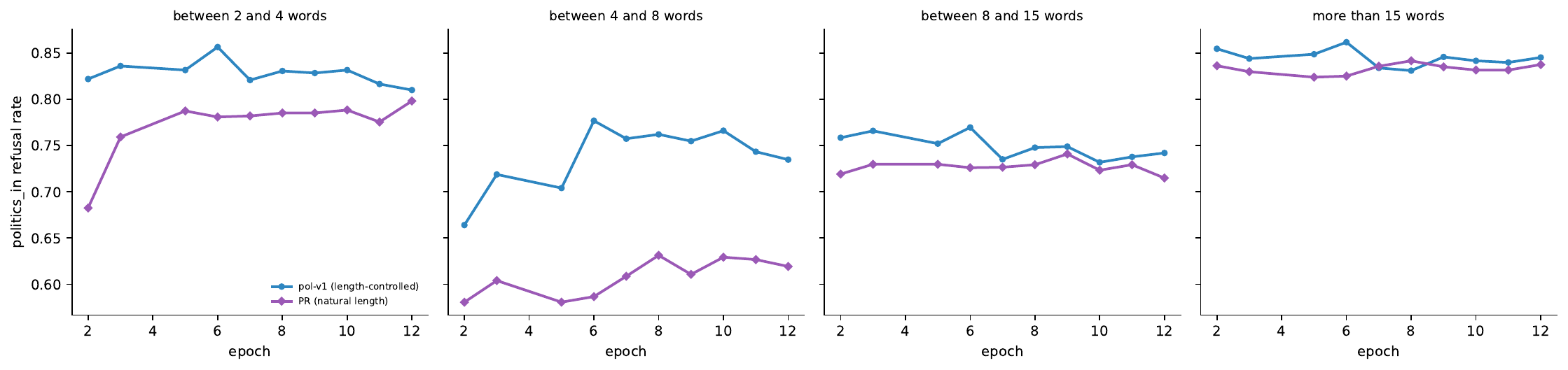}
  \caption{\textbf{Refusal rate by prompt length.} Length-controlled CE generation is
  compared with pairwise-generated PR across four prompt-length buckets.}
  \label{fig:length_analysis}
\end{figure*}

The comparison uses the v1 CE run and the PR-only run at matched evaluated epochs. The
length-controlled run reaches a higher political refusal rate in every bucket. The difference
is largest for short prompts and narrows for prompts longer than 15 words, where the natural PR
distribution has stronger coverage. This result supports length control as a secondary data
construction choice rather than as a separate safety mechanism.

\subsection{Pairwise Boundary Data Without SC}
\label{app:a1downside}

\begin{figure}[t]
  \centering
  \includegraphics[width=\linewidth]{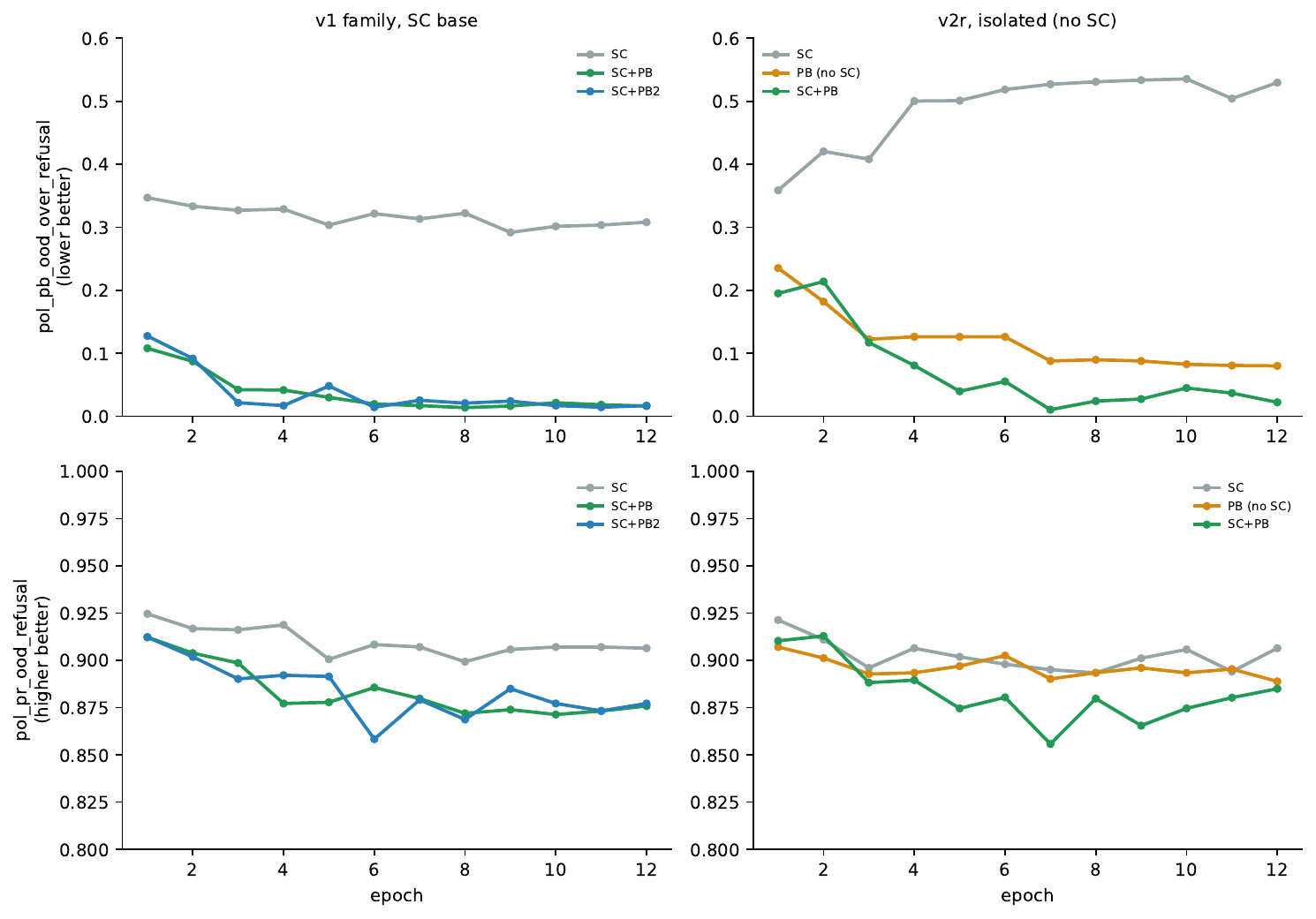}
  \caption{%
    \textbf{Pairwise boundary data reduces false-positive refusals at the boundary.}
    Adding PB lowers \texttt{pol\_pb\_ood\_over\_refusal} sharply both with an SC
    base (left, v1 family) and in isolation without SC (right, v2r), so the
    effect does not depend on SC.
    \label{fig:downside_ablation_pb_sc}
    Verification on top of PB (PB2, left) gives no significant further
    improvement over PB alone.
    \label{fig:downside_ablation_pb_isolated}
    Combining SC with PB gives a slight additional gain over PB without SC
    (right). \texttt{pol\_pr\_ood\_refusal} stays high throughout.
  }
\end{figure}

\section{Qualitative Analysis and Data Inspection}

\subsection{Component Overlap and Response Regeneration}
\label{app:qual_promptresponse}

\begin{table}[t]
\centering
\small
\caption{Prompt overlap between related benign components. Counts use exact prompt-text
matching in the audited files.}
\label{tab:component-overlap}
\footnotesize
\setlength{\tabcolsep}{3pt}
\begin{tabular}{lrrr}
\toprule
\textbf{Comparison} & \textbf{First} & \textbf{Second} & \textbf{Shared} \\
\midrule
PB vs PB2 & 4,986 & 5,000 & 4,961 \\
SC vs SC2 & 5,000 & 4,954 & 4,953 \\
\bottomrule
\end{tabular}
\end{table}

PB and PB2 draw from the same pairwise comply-side source, but they are separate generated
files and are not text-identical copies. Their overlap confirms that the comparison is largely
prompt matched while also exposing a small construction difference. SC2 preserves almost the
entire SafeChain prompt set while replacing the adopted response with a target-model response.
In the sampled matched rows reported by the audit, the paired responses were not identical in
either comparison.

\subsection{Boundary Pair Examples}
\label{app:qual_boundarypairs}

The two examples in \cref{tab:boundary-examples} preserve a named event or institution while
changing the requested behaviour. The benign prompt asks for factual explanation. The target
harmful prompt asks the model to advocate or persuade. This demonstrates that the evaluation is
not based only on different topic vocabulary.

\subsection{Representative FakeHarm Grid Cells}
\label{app:qual_fakeharm}

Representative generated cells include \emph{shoot} in the context of competitive games,
\emph{morphine} in an academic chemistry context, \emph{attack} in electoral campaign
analysis, \emph{targeted killing} in an international-relations paper, and \emph{liquidate
assets} in a venture-capital context. These cells illustrate how a dangerous surface cue is
paired with a context that establishes the benign interpretation. Full prompt text is not
reported because the audit exported only focus and context metadata for these examples.

\subsection{Manual Inspection Procedure}
\label{app:qual_manualinspection}

Qualitative inspection uses \texttt{tools/manual\_eval\_inspect.py}. The tool samples a fixed
number of examples per evaluation dataset, ten by default, with a fixed seed. It reuses an
existing response when available and generates a new response only when the saved evaluation
output is absent. Each sampled response is then scored with the judge used for the corresponding
metric. The output contains a structured JSON file and a readable text report with the prompt,
response, judge decision, and available metadata.

This procedure is a researcher-operated diagnostic check. It is not a multi-annotator study.
No independent reviewer protocol, agreement statistic, or disagreement-resolution procedure
was used.

\subsection{Observed Construction Artefacts}
\label{app:qual_artefacts}

The PB production file predates response verification. PB2 should therefore be treated as a
separate verified rebuild rather than as a filtered view of the exact PB rows. The current
repair code also shows that both Graft variants sample with replacement. Any earlier explanation
based on different replacement rules is superseded.

The audited v3 and v3r files are not row aligned. v3 contains 40,401 retained rows and 79
residual failures, while v3r contains 40,293 rows. This prevents a row-level qualitative
comparison between the two artifacts without recovering their original build provenance.

\section{Additional Completed Ablation}

\subsection{FakeHarm and Boundary Data Interaction}
\label{app:fhpbablation}

Adding FakeHarm to a mixture that already contains PB further reduces XSTest over-refusal, as
shown in the fourth panel of \cref{fig:fakeharm_family} (\cref{fig:fakeharm_pb_interaction}).
This supports the interpretation that the two components target different benign regions. PB
improves local refusal-boundary precision, while FakeHarm provides calibration for benign
prompts with dangerous surface cues.

\subsection{Religion as a Second Constructed Topic}
\label{app:religion_scope}

The data-generation pipeline described in \cref{sec:method} is topic-agnostic and also supports
religion as a domain. Using the same four-stage hierarchical construction and
persona-conditioned generation, we built a religion-domain prompt set with 30 personas across
five ideological clusters (manipulative, sectarian, anti-religion, ideological, and other
radical framings), mirroring the political persona pool in structure. The construction includes
the main refusal data and paired harmful and benign boundary components. The religion and
politics datasets will be released together on Hugging Face after the remaining documentation
and release checks are complete. Religion-domain training artifacts exist, but the required
religion-specific evaluation was not completed for this paper. Our experimental claims are
therefore restricted to political persuasion, and no religion-domain result is claimed or
implied here.

\end{document}